%% file: main.tex
\documentclass[sigconf]{acmart}
\AtBeginDocument{%
  }

\usepackage{pifont}
\usepackage{colortbl}
\usepackage{multirow}
\usepackage{url}

\usepackage{amsmath,amssymb,amsfonts, bm}
\usepackage{colortbl} 
\usepackage{xcolor}
\usepackage{array}  
\usepackage{subfigure}
\setcopyright{acmcopyright}
\usepackage{twemojis}
\newtheorem{theorem}{Theorem}

\newtheorem{lemma}{Lemma}
\usepackage[ruled,linesnumbered]{algorithm2e}
\usepackage[export]{adjustbox}
\usepackage{multicol}
\usepackage{enumitem}
\usepackage{titletoc}
\usepackage{tcolorbox}
\usepackage{hyperref}
\hypersetup{
    colorlinks=True,
    linkcolor=blue,
    citecolor=blue,
    urlcolor=black
}
\SetKwRepeat{Do}{do}{while}

\setcopyright{acmlicensed}
\copyrightyear{2025}
\acmYear{2025}
\setcopyright{acmlicensed}\acmConference[KDD '25]{Proceedings of the 31st ACM SIGKDD Conference on Knowledge Discovery and Data Mining V.2}{August 3--7, 2025}{Toronto, ON, Canada}
\acmBooktitle{Proceedings of the 31st ACM SIGKDD Conference on Knowledge Discovery and Data Mining V.2 (KDD '25), August 3--7, 2025, Toronto, ON, Canada}
\acmDOI{10.1145/3711896.3736823}
\acmISBN{979-8-4007-1454-2/2025/08}

\begin{document}

\title{A Two-Stage Data Selection Framework for Data-Efficient Model Training on Edge Devices}

\author{Chen Gong}
\affiliation{
    \institution{Shanghai Jiao Tong University}
    \city{Shanghai}
    \country{China}
}
\email{gongchen@sjtu.edu.cn}
\author{Rui Xing}
\affiliation{
    \institution{Shanghai Jiao Tong University}
    \city{Shanghai}
    \country{China}
}
\email{cocojess@sjtu.edu.cn}
\author{Zhenzhe Zheng}
\affiliation{
    \institution{Shanghai Jiao Tong University}
    \city{Shanghai}
    \country{China}
    \authornote{Zhenzhe Zheng is the corresponding author.}
}
\email{zhengzhenzhe@sjtu.edu.cn}
\author{Fan Wu}
\affiliation{
    \institution{Shanghai Jiao Tong University}
    \city{Shanghai}
    \country{China}
}
\email{fwu@cs.sjtu.edu.cn}


\input{Contents/0-Abstract}

\ccsdesc[500]{Human-centered computing~Mobile computing}
\ccsdesc[500]{Computing methodologies~Machine learning}

\keywords{On-Device Machine Learning, Data Selection and Utilization}

\maketitle

\input{Contents/1-Introduction}
\input{Contents/2-Background}
\input{Contents/3-Design}
\input{Contents/4-Evaluation}

\input{Contents/5-RelatedWork}

\input{Contents/6-Conclusion}

\begin{acks}
    This work was supported in part by National Key R\&D Program of China (No. 2023YFB4502400), in part by China NSF grant No. 62322206, 62025204, 62132018, U2268204, 62272307, 62372296. The opinions, findings, conclusions, and recommendations expressed in this paper are those of the authors and do not necessarily reflect the views of the funding agencies or the government. The authors thank the anonymous reviewers for their insightful feedbacks.
\end{acks}
\clearpage
\bibliographystyle{ACM-Reference-Format}
\bibliography{main.bib}
\input{Contents/7-Appendix}

\end{document}

%% file: Contents/0-Abstract.tex
\begin{abstract}
The demand for machine learning (ML) model training on edge devices is escalating due to data privacy and personalized service needs. 
However, we observe that current on-device model training is hampered by the under-utilization of on-device data, due to low training throughput, limited storage and diverse data importance.
To improve data resource utilization, we propose a two-stage data selection framework {\sf Titan} to select the most important data batch from streaming data for model training with guaranteed efficiency and effectiveness.
Specifically, in the first stage, {\sf Titan} filters out a candidate dataset with potentially high importance in a coarse-grained manner.
In the second stage of fine-grained selection, we propose a theoretically optimal data selection strategy to identify the data batch with the highest model performance improvement to current training round.
To further enhance time-and-resource efficiency, {\sf Titan} leverages a pipeline to co-execute data selection and model training, and avoids resource conflicts by exploiting idle computing resources.
We evaluate {\sf Titan} on real-world edge devices and three representative edge computing tasks with diverse models and data modalities. Empirical results demonstrate that {\sf Titan} achieves up to $43\%$ reduction in training time and $6.2\%$ increase in final accuracy with minor system overhead, such as data processing delay, memory footprint and energy consumption. 
\end{abstract}

%% file: Contents/1-Introduction.tex
\section{Introduction}
\label{sec: introduction}
\textit{Machine learning (ML)} models have been widely embedded in mobile applications to provide diverse intelligent services, such as image tagging in Google Lens~\cite{google_lens}, command recognition in Siri~\cite{siri}, text prediction in Microsoft SwiftKey~\cite{keyboard}, etc.
With growing concerns over data privacy and higher demands on personalized model performance
~\cite{sun2025tensorshieldsafeguardingondeviceinference, TP-toolbox-web}, on-device model training is becoming necessary to facilitate 
a single device to adapt the local model to its own data distribution~\cite{DBLP:conf/mobisys/GimK22, gong2024delta, xu2019first, sk2022characterizing, DBLP:conf/www/Liu0ZL025, zhuang2024litemoe}, or 
multiple devices to collaboratively train a global model that can generalize well across different data distributions~\cite{DBLP:conf/aistats/McMahanMRHA17, gong2023store, gong2025ode}. 

The success of ML models highly relies on the abundant valuable data for model training~\cite{DBLP:conf/icml/GhorbaniZ19, DBLP:conf/iccv/SunSSG17,goyal2017accurate}. 
On one hand, a large-scale training dataset is essential to the generalization performance of ML models~\cite{DBLP:conf/iccv/SunSSG17}. 
On the other hand, massive data aids in stabilizing the model training process and reducing the training time to reach a target accuracy~\cite{goyal2017accurate}. 
As a result, on cloud server side, it is common to collect extensive training data for iterative model updates, such as 29 million games for training AlphaGo~\cite{DBLP:journals/nature/SilverSSAHGHBLB17} and $500$ billion tokens for pre-training ChatGPT3~\cite{chat-gpt}. 
Similarly, on mobile devices side, it is also desirable to fully utilize the on-device sensor data for model training to achieve a satisfactory model training performance and improve user experience.


\textbf{Motivation.} 
Previous works for on-device model training mainly focused on the exploitation of limited \textit{hardware resources}, such as
optimizing memory allocation to increase batch size during training~\cite{DBLP:conf/mobisys/WangXJDY0HLL22, DBLP:conf/mobisys/GimK22}, 
co-using multiple types of computing resources to speed up model inference and training~\cite{DBLP:conf/mobicom/XuXWW0H0JL22, DBLP:conf/mobisys/JiaZCJLRZ22, DBLP:conf/mobicom/WangDCLX21, tan2022deep}, 
dynamically adjusting the size of trainable parameters to improve training efficiency~\cite{jia2023elastictrainer,wei2023nn}, etc. 
However, we observe that the under-utilization of \textit{data resource} is another key bottleneck to on-device model training, 
which results in up to $3.5\times$ longer training time and $13.3\%$ lower final accuracy in our pilot experiments due to low training throughput, limited storage and diverse data importance (elaborated in \S\ref{sec: Resource Limitation Enforces On-Device Data Selection}).
Therefore, a crucial open problem is: \textit{Is it possible to design an on-device data selection framework to concentrate the limited hardware resources on important training data for superior model training performance?}

\textbf{Challenges.}
The design of an on-device data selection framework needs to achieve \textit{effectiveness} and \textit{efficiency} simultaneously:\\
\textit{i) Effectiveness:} As the on-device model performance directly impacts the quality of application service and user experience, it is necessary for the data selection framework to provide theoretical and empirical guarantees on enhancing model training performance.
\textit{ii) Efficiency:} For deployment, the data selection framework is desired to be time-and-resource efficient.
First, the application data typically undertakes real-time services like teleconferencing, which requires the data selection process to be low-latency to avoid compromising user experience.
Second, the data selection framework needs to avoid intense resource conflict with model training process, which would extend the time of each model update and offset the performance improvement brought by data selection.

It is challenging to satisfy these two properties simultaneously.
Higher effectiveness necessitates a more accurate but time-intensive data importance evaluation process over a broader candidate dataset, which inevitably increases per-sample latency and consumes more computing resources.
Our theoretical analysis and experimental results in \S\ref{sec: limitation of state-of-the-art} indicate that conventional cloud-side data selection approaches such as importance sampling~\cite{DBLP:conf/icml/KatharopoulosF18, DBLP:conf/icml/ZhaoZ15}, heuristic selection~\cite{DBLP:conf/iclr/ColemanYMMBLLZ20, DBLP:conf/cvpr/RebuffiKSL17, settles2009active, DBLP:conf/iclr/YoonMYH22, everaert2023gio} and coreset selection~\cite{DBLP:conf/icml/MirzasoleimanBL20, pooladzandi2022adaptive, DBLP:conf/sigmod/LiSC22} fail to be applied to device side due to ineffectiveness or inefficiency.

\textbf{Our Solutions}.
In this work, we address the above challenge by proposing a \underline{t}wo-stage onl\underline{i}ne da\underline{ta} selectio\underline{n} framework {\sf Titan}, which simultaneously achieves high effectiveness and efficiency for on-device data utilization.
First, to guarantee the effectiveness, we theoretically analyze the correlation between the training data batch and the on-device model training performance, based on which we demonstrate the sub-optimality of the state-of-the-art importance sampling approach due to overlooking a crucial term of class variance during inter-class batch size allocation.
Further, we propose a \textit{theoretically optimal data selection strategy} to identify the data batch with the highest improvement to model performance in each training round.
Second, to improve time-efficiency, {\sffamily Titan} employs a \textit{two-stage architecture}.
In the first stage, {\sffamily Titan} leverages a carefully designed coarse-grained filter to estimate the potential importance of each streaming data sample within millisecond-level latency, and locally buffers a small candidate dataset. 
In the second stage, the buffered candidate dataset undergoes our proposed data selection strategy to enhance effectiveness.
Third, for higher time-and-resource efficiency, we design a \textit{pipeline} to facilitate the co-execution of model training and data selection, and exploits the idle computing resources commonly seen on devices to mitigate potential resource conflicts.


\textbf{Contributions} of this work are summarized as follows:
\begin{itemize}[leftmargin=0.3cm, topsep=0cm]
    \item To the best of our knowledge, we are the first to point out the severity of data resource under-utilization in on-device model training process, and conduct in-depth analysis for this issue.
    \item We perform comprehensive evaluation of existing cloud-side data selection approaches for device-side setting, and provide theoretical and empirical analysis on their failures.
    \item We propose an on-device data selection framework  {\sf Titan}, consisting of a theoretically optimal data selection strategy, a two-stage architecture and a pipeline design, to simultaneously achieve high efficiency and effectiveness for on-device data utilization.
    \item We implement {\sffamily Titan} framework on real-world device and demonstrate {\sf Titan}'s superiority across three typical mobile computing tasks with varied data modalities and ML models.
\end{itemize}

%% file: Contents/2-Background.tex
\section{Background and Motivation}
\label{Preliminaries}
In this section, we first briefly introduce the background of on-device model training (\S\ref{sec: on-device model training}). Then, we delve into the under-utilization of on-device data resources to illustrate the motivation of this work (\S\ref{sec: Resource Limitation Enforces On-Device Data Selection}). Next, we elaborate the limitations of existing cloud-side data selection methods (\S\ref{sec: limitation of state-of-the-art}). 

\subsection{On-Device Model Training}
\label{sec: on-device model training}
Similar to cloud-side ML, the objective of on-device model training can be formulated as minimizing the loss function $L(w,\mathcal{P})$, which represents the prediction error (or loss) of model with parameters $w$ on local data distribution $\mathcal{P}$:
\begin{equation}
        w^*=\min_w L(w, \mathcal{P})\overset{\text{def}}{=}\mathbb{E}_{(x,y)\sim \mathcal{P}}[l(w,x,y)],
    \label{eq: model update}
\end{equation}
where $\mathbb{E}_{(x,y)\sim\mathcal{P}}\big[l(w,x,y)\big]$ denotes the expected loss (or error) of model $w$ over 
data $(x,y)$ following distribution $\mathcal{P}$. 

In on-device model training, mini-batch SGD~\cite{robbins1951stochastic} is widely adopted to solve the above optimization problem (\ref{eq: model update}), which involves three steps in each training round $t$:

\noindent \textit{1) Data Collection}: 
    Data samples $(x,y)\!\sim\!\mathcal{P}$ are continuously collected by device in a streaming manner and stored in the local storage. We use $\mathcal{S}$ and $\mathcal{S}_y$ to denote the sets of all the stored data samples and the data samples with class $y\!\in\!\mathcal{Y}$.

\noindent \textit{2) Data Loading}: 
    A batch of data samples {\small $\mathcal{B}\!=\!\{(x_i,y_i)|1\!\le\! i\!\le\! |\mathcal{B}|\}$} is loaded from storage to memory as training data.

\noindent \textit{3) Model Update}:
    Current model parameters $w_{t}$ are updated by the average gradient of the loaded training data batch:
    \begin{equation}
            w_{t+1}= w_{t}-\eta_t\cdot \mathbb{E}_{(x,y)\in \mathcal{B}}\big[\nabla_w l(w_{t},x,y)\big],
        \nonumber
    \end{equation}
    where $w_t$ denotes the updated model parameter in training round $t$ and $\eta_t$ is the corresponding learning rate.
    
Typically, data collection is conducted concurrently with data loading and model update, both of which are executed iteratively.

\subsection{Under-Utilization of On-Device Data}
\label{sec: Resource Limitation Enforces On-Device Data Selection}
We elaborate the three unique characteristics of mobile devices, which leads to the under-utilization of on-device data and motivates the design of a device-specific data selection framework.

\textbf{Low data throughput during model training.} 
The limited memory and computing resources of devices restrict the training data throughput. 
First, the memory size constrains the number of data samples that can be co-processed within a data batch (\textit{e.g.} batch size $16$ for common lightweight model MobileNetV1 has reached the limit of high-end devices like MI 9 with $6$GB RAM~\cite{cai2021towards}).
Second, the on-device per-sample training time is relatively long due to the limited computing hardware~\cite{DBLP:conf/mobisys/JiaZCJLRZ22}. Specifically, the forward-and-backward propagation over modern ML models can be time-consuming (\textit{e.g.} it takes around $20$s for the representative device Jetson Nano to train one data batch with size $16$ on MobileNetV1).

\textbf{Limited on-device storage for training data.}  
In numerous mobile applications, on-device data is continuously collected in a streaming manner, but devices typically have quite limited storage for training data due to user preference as well as software and hardware limitation. 
On one hand, users usually prioritize reserving storage space for personal files like photos, documents and chat history instead of each application's training data.
On the other hand, both iOS and Android platforms impose size limitations on applications, such as less than 4GB for an iOS app~\cite{ios_space} and no more than 2GB for a Google Play app~\cite{android_space}.
Furthermore, for low-end devices like HUAWEI WiFi AX3~\cite{huawei_router} with less than 1GB storage, it is impractical to save all the collected data for model training.

\textbf{Diverse data importance to training performance.} 
The on-device data samples have diverse importance (or quality) for model training, stemming from 
\textit{1)} the wide range of on-device data distribution caused by varied user behaviors and application services at different times of a day,
and \textit{2)} heterogeneous data quality due to sensors from different producers and unstable network environments.
Consequently, the involvement of low-importance data in model training will further reduce the on-device data utilization.

\begin{figure}
    \centering
    \includegraphics[width=0.48\textwidth]{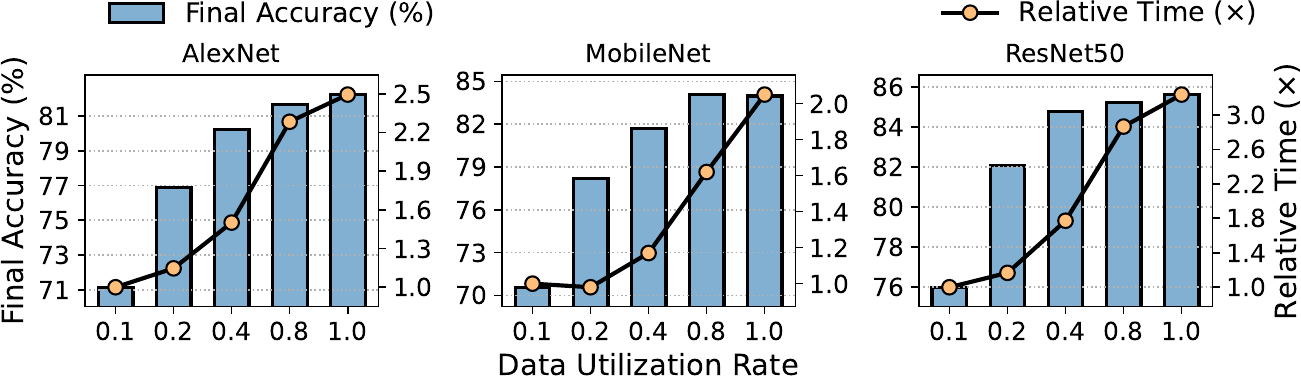}
    \vspace{-0.2cm}
    \caption{Final inference accuracy and normalized training time of different models and data utilization rates.}
    \Description{Final inference accuracy and normalized training time of different models and data utilization rates.}
    \vspace{-0.2cm}
    \label{fig: the impact of data resource utilization on model training}
\end{figure}

\textbf{Motivating Experiments.}
The aforementioned properties restrict the on-device data utilization and hinder the success of on-device model training processes.
On one hand, if we attempt to utilize all the collected data for parameter update to achieve superior training performance, all the samples need to be incorporated in each training round, which leads to substantial per-round training time and storage overhead~\cite{DBLP:conf/iclr/SmithKYL18,DBLP:conf/mobisys/WangXJDY0HLL22}. 
On the other hand, if we leverage only partial data for higher efficiency of model training and data storage, the parameter update computed from partial data tend to deviate from the expected update computed from all data, thereby degrading the final model accuracy~\cite{DBLP:conf/icml/KatharopoulosF18}. 
Our preliminary experiments on representative device Jetson Nano and dataset CIFAR-10~\cite{krizhevsky2009learning} in Figure \ref{fig: the impact of data resource utilization on model training} show that leveraging only partial data resource can reduce the final accuracy by $9.6\!-\!13.4\%$ while the utilization of full data will prolong total training time by $2.05\!-\!3.24\times$.
Therefore, an on-device data selection framework is necessary to focus limited hardware resources on partial but important data resources for higher data utilization efficiency.

\subsection{Existing Data Selection Approaches}
\label{sec: limitation of state-of-the-art}
Existing cloud-side data selection approaches can hardly be applied to the device side due to ineffectiveness or inefficiency.

\textbf{Importance sampling} ({\sf IS})~\cite{DBLP:conf/icml/KatharopoulosF18, DBLP:conf/icml/ZhaoZ15} is the state-of-the-art data selection approach, which selects each training data sample according to its importance to model training performance. 
Previous research has demonstrated a negative correlation between the gradient variance of the training data batch\footnote{The variance represents the average difference between the gradient of the selected training data and the expected gradient of the entire dataset.} and the model training performance. 
As a result, {\sffamily IS} defines the per-sample importance as its gradient norm to minimize such gradient variance and optimize the training performance.

\begin{figure}
    \subfigure[Per-round training time.]{
        \includegraphics[height=2.4cm]{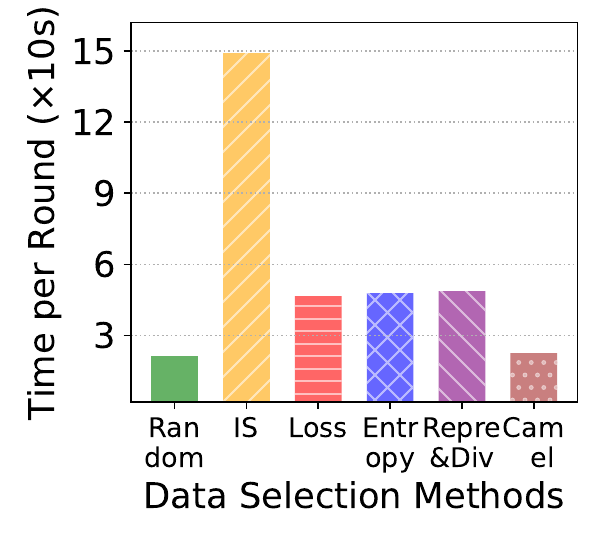}
        \label{fig: limitation of data selection time}
    }
    \ 
    \subfigure[Training processes with batch sizes 10 and 25.]{
        \includegraphics[height=2.4cm]{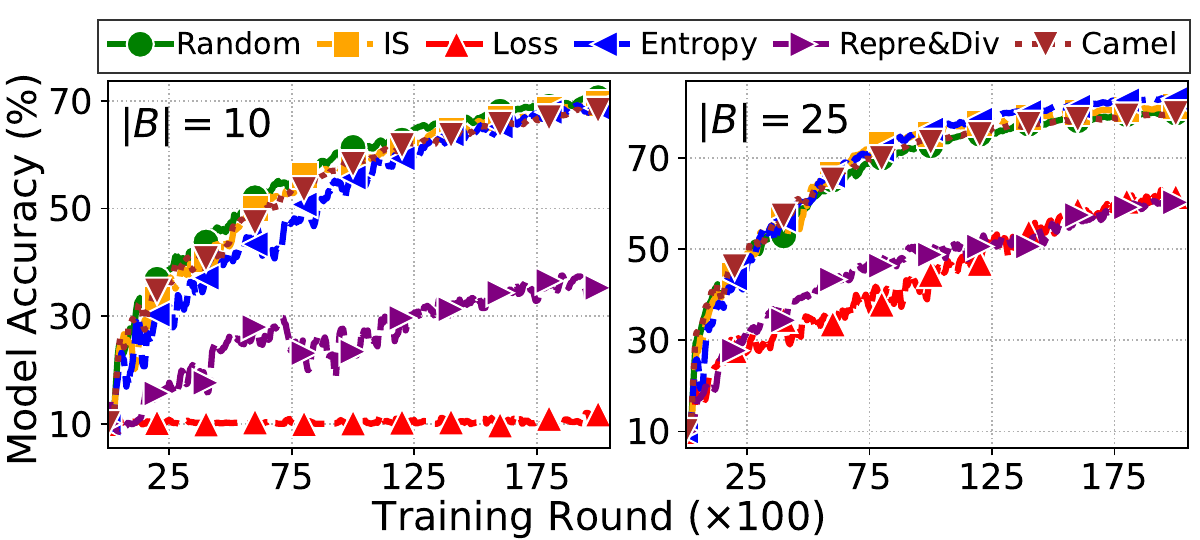}
        \label{fig: limitation of data selection}
    }
    \vspace{-0.2cm}
    \caption{Per-round training time and training curves of different data selection methods on MobileNetV1 and CIFAR-10, tested on real device Jetson Nano.}
    \Description{Per-round training time and training curves of different data selection methods on MobileNetV1 and CIFAR-10, tested on real device Jetson Nano.}
    \vspace{-0.2cm}
\end{figure}

However, {\sffamily IS} is neither effective nor efficient for devices. 
On one hand, we identify that the theoretical optimality of {\sffamily IS} relies on an underlying assumption that each sample in the training data batch is independently selected, which leads to sub-optimal performance for batch-level selection, especially for small training batches on devices.
The detailed theoretical analysis and verification results are provided in \S\ref{sec: fine-grained data selection}.
On the other hand, {\sffamily IS} requires computing each sample's gradient over model parameters, which can prolong the per-round training time by up to $7\times$ shown in Figure \ref{fig: limitation of data selection time}.
For device-side efficient deployment, Mercury~\cite{zeng2021mercury} proposed to divide the dataset into multiple subsets and recompute only one subset's importance per training round, which however, is not applicable for real-world mobile computing tasks involving streaming data.

\textbf{Heuristic data selection} ({\sffamily HDS}) enhances model training efficiency by selecting the training data with various intuitive metrics, such as 
model uncertainty quantified by loss or entropy of model output logits~\cite{DBLP:conf/iclr/ColemanYMMBLLZ20, settles2009active}, 
data representativeness measured by closeness to the distribution centroid in feature space~\cite{everaert2023gio} and diversity to other samples~\cite{DBLP:conf/iclr/YoonMYH22}, etc. 

We find that {\sffamily HDS} is efficient but lacks effectiveness from both theoretical and empirical aspects.
Theoretically, existing {\sffamily HDS} fails to directly correlate the data importance metric with model training performance, thereby essentially optimizing a proxy objective of intuitively defined metrics rather than the fundamental objective (\ref{eq: model update}) of model training performance.
Therefore, practical implementation of {\sffamily HDS} often involves cumbersome trial-and-error processes to explore the appropriate metrics that could bring the highest model performance improvement. 
Empirically, Figures \ref{fig: limitation of data selection} reveals that {\sffamily HDS} (\textit{i.e.} \textit{Loss}, \textit{Entropy} and \textit{Repre\&Div}) even leads to degraded training performance compared with random selection when the batch size is small. 
This is because traditional {\sffamily HDS} relies on large batch sizes to mitigate the distribution deviation and parameter update bias of the heuristically selected training data batch.

\begin{figure*}
    \begin{minipage}[b]{0.65\textwidth}
        \centering
        \includegraphics[width=\textwidth]{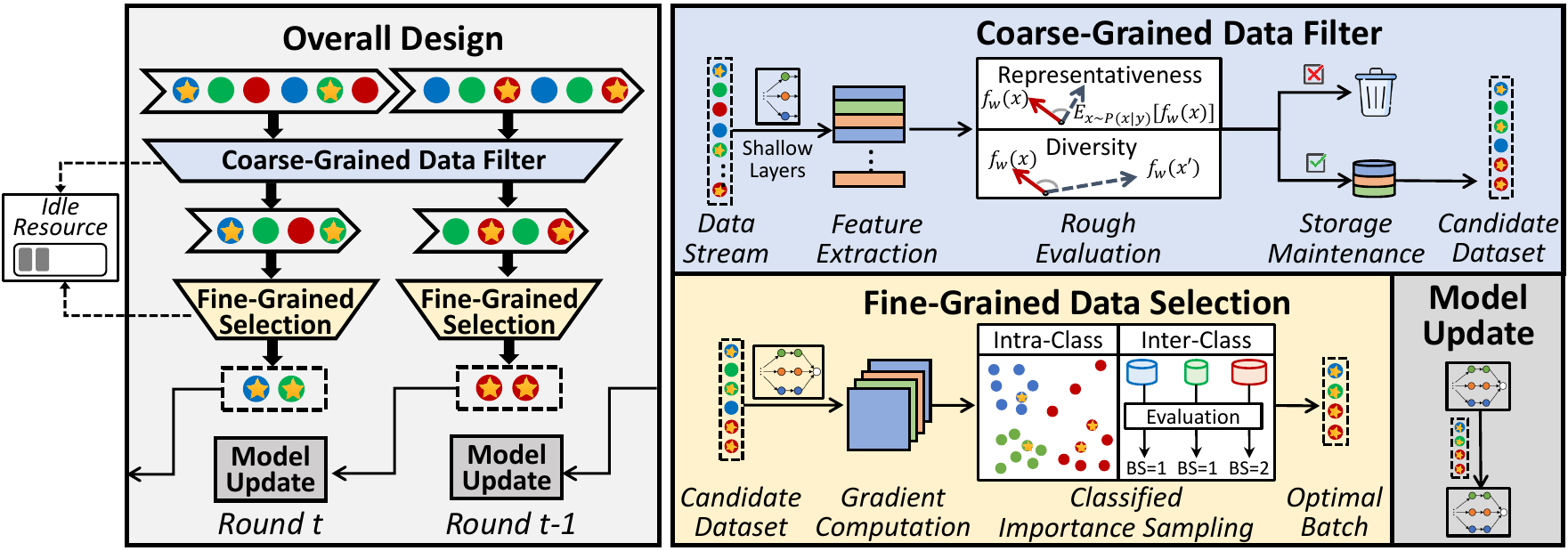}
        \vspace{-0.5cm}
        \caption{Overall design and workflow of {\sf Titan}.}
        \Description{Overall design and workflow of {\sf Titan}.}
        \label{fig: overview}
    \end{minipage}
    \ \ 
    \begin{minipage}[b]{0.33\textwidth}
        \centering
        \includegraphics[width=0.7\textwidth]{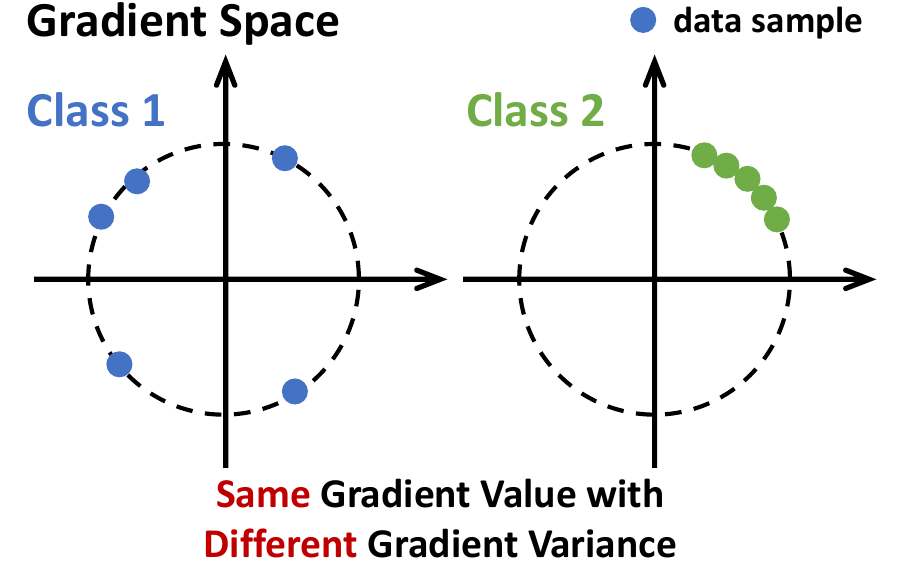}
        \vspace{-0.3cm}
        \caption{Comparison between {\sf IS} and {\sf C-IS}. {\sf IS} will select an equal number of data samples from classes 1 and 2 , while {\sf C-IS} will select more samples from class 1 by considering its larger gradient variance.}
        \Description{Comparison between {\sf IS} and {\sf C-IS}. {\sf IS} will select an equal number of data samples from classes 1 and 2 , while {\sf C-IS} will select more samples from class 1 by considering its larger gradient variance.}
        \label{fig: simple example}
    \end{minipage}
\end{figure*}

\textbf{Coreset Selection} ({\sffamily CS})~\cite{DBLP:conf/icml/MirzasoleimanBL20, pooladzandi2022adaptive, DBLP:conf/sigmod/LiSC22} aims to select a small weighted data subset, \textit{i.e.} coreset, to approximate the entire dataset in terms of gradient computation, thereby reducing the training data scale without significant deviation in parameter update direction. Previous research formulated the gradient estimation error of the coreset as a sub-modular function, and derived the optimal coreset by minimizing such error.

We observe that {\sffamily CS} is either inefficient or ineffective for devices. 
To select a coreset with size $|\mathcal{B}|$ from $|\mathcal{S}|$ data samples, {\sffamily CS} requires computing the gradients of all $|\mathcal{S}|$ samples to solve the error minimization problem, incurring high computation overhead similar to {\sffamily IS}. 
Instead of directly minimizing the gradient distance between coreset and the entire dataset, \textit{Camel}~\cite{DBLP:conf/sigmod/LiSC22} uppers bound the gradient distance by raw input distance to avoid the cumbersome model backpropagation process. However, this approximation compromises the theoretical guarantee of {\sffamily CS} and also exhibits inferior performance in our prior experiments in Figure \ref{fig: limitation of data selection}. This is because the complex structures of modern ML models and the wide distribution of on-device data make the raw data distance unable to accurately reflect the gradient distance.


%% file: Contents/3-Design.tex
\section{Design of Titan}
\label{section: design}
In this section, we first give an overview of {\sffamily Titan} framework (\S\ref{sec: overview}), and then we elaborate each key component in {\sf Titan} (\S\ref{sec: fine-grained data selection}-\S\ref{sec: pipeline design}).

\subsection{Overview}
\label{sec: overview}
\textbf{Design Rationale.}{\sffamily Titan} aims to exploit on-device data resources effectively and efficiently by incorporating three key designs:
\textit{1) a theoretically optimal strategy} for training data batch selection, which operates as a fine-grained selection component to identify the data batch that brings the highest improvement to model performance,
\textit{2) a coarse-grained filter} to filter out a small candidate dataset from streaming data in real time through heuristic metrics specially co-designed with the optimal fine-grained selection component,
\textit{3) a pipeline design} to co-execute the processes of data selection and model training and mitigate their potential resource conflict by utilizing on-device idle computing resources.

\textbf{Workflow.}
As depicted in Figure \ref{fig: overview}, {\sf Titan} adopts a two-stage architecture and forks three concurrent processes to steadily select the optimal training data batch from real-time data streams for each round of model training: \\
\textit{1) Coarse-grained filter}:
    Whenever a data sample is collected by device, {\sffamily Titan} extracts its feature by inputting the data into shallow layers of ML models, and estimates its potential importance within milliseconds through two specially designed heuristic metrics. 
    Then, {\sffamily Titan} maintains a candidate dataset in local buffer with a priority queue to facilitate the subsequent fine-grained selection.\\
\textit{2) Fine-grained selection}:
    During each round $t$, {\sffamily Titan} computes the gradient for each buffered data sample over the final model layer, and identifies the ideal data batch for next round $t+1$ through the proposed optimal data selection strategy.\\
\textit{3) Model update}:
    Simultaneously in round $t$, current model parameter $w_t$ is updated by the data batch chosen in preceding round $t-1$, which forms a seamless pipeline with the fine-grained selection process for the upcoming round.


\subsection{Fine-Grained Data Selection}
\label{sec: fine-grained data selection}

To optimize the effectiveness of on-device data selection, we propose a new data batch selection strategy for mini-batch SGD, namely classified importance sampling ({\sf C-IS}), which consists of inter-class batch size allocation and intra-class data selection.
We first provide the definitions of \textit{class importance} and \textit{sample importance}, which are used to determine \textit{how many} and \textit{which} data samples to select for each class. 
Then, we analyze {\sf C-IS}'s theoretical optimality in improving on-device model training performance and provide an intuitive explanation for better understanding.

\textit{Inter-Class Batch Size Allocation.}
To select a batch of training data with size $|\mathcal{B}|$, {\sf C-IS} determines the data selection size $|\mathcal{B}_y|$ for each class $y\!\in\!\mathcal{Y}$ according to the class importance $I_t(y)$ in current round $t$, which is defined as:
\begin{equation}
    \begin{small}
    \begin{aligned}
        & I_t(y) \overset{\text{def}}{=} \\
        & |\mathcal{S}_y|\Big[\underbrace{\mathbb{V}_{(x,y)\sim P_{t, y}}\!\big[\nabla l(w_t,x,y)\big]}_\mathrm{variance\ of\ gradient}\!-\!\underbrace{\mathbb{V}_{(x,y)\sim P_{t, y}}\!\big[||\nabla l(w_t,x,y)||_2\big]}_\mathrm{variance\ of\ gradient\ norm}\Big]^{\frac{1}{2}}
    \end{aligned}
    \label{eq: class importance}
    \end{small}
\end{equation}
where $|\mathcal{S}_y|$ denotes the total amount of stored data samples with class $y$ and $\mathbb{V}_{(x,y)\sim P_{t,y}}[f(x)]$ denotes the variance of function $f(x)$ with selection probability $P_{t,y}(x)$ for each data sample $x$ in class $y$ and $||*||_2$ denotes the $l_2$-norm. 

\textit{Intra-Class Data Selection.} 
To select $|\mathcal{B}_y|$ important data samples from the stored data $\mathcal{S}_y$ of class $y\!\in\!\mathcal{Y}$, we select each sample $(x,y)$ with probability
proportional to its sample importance $I_t(x,y)$, which is defined as:
\begin{equation}
    I_t(x,y)\triangleq\underbrace{\big|\big|\nabla l(w_t,x,y)\big|\big|_2}_\mathrm{gradient\ norm}.
    \label{eq: sample importance}
\end{equation}

\textbf{Theoretical Analysis.}
To demonstrate the theoretical optimality of {\sf C-IS}, we first
present Theorem \ref{theorem: convergence rate} and Lemma \ref{lemma: optimal distribution for sgd} to analyze the impact of training data batch on model training performance as well as the sub-optimality of state-of-the-art {\sf IS} in data batch selection.
Further, we present Theorem \ref{theorem: variance decomposition for mini-batch sgd} and Lemma \ref{lemma: optimal distribution for minibatch sgd} to demonstrate the optimality of our proposed {\sf C-IS} in improving model training performance. 

Previous studies~\cite{DBLP:conf/icml/ZhaoZ15,  DBLP:conf/icml/KatharopoulosF18} have demonstrated a negative correlation between the gradient variance of the training data batch $\mathcal{B}$ and model training performance, where the performance of model with parameter $w$ is quantified by its distance to the optimal parameter $w^*$ (\textit{i.e.} $||w-w^*||_2^2$). Therefore, the model training performance in round $t$ can be measured by the decrease in the distance to $w^*$ from initial model parameter $w_{t}$ to updated model parameter $w_{t+1}$:
\begin{theorem}[Training Performance Measurement~\cite{DBLP:conf/icml/ZhaoZ15,  DBLP:conf/icml/KatharopoulosF18}]
    \label{theorem: convergence rate}
    The model training performance in round $t$ is negatively correlated with the gradient variance of the training data batch $\mathcal{B}$ selected by data selection strategy $P_t$:
    \begin{equation}
        \begin{small}
        \begin{aligned}
            & \mathbb{E}_{\mathcal{B}\sim P_t}\Big[\underbrace{|| w_t\!-\!w^*||_2^2-|| w_{t+1}\!-\!w^*||_2^2}_{\mathrm{reduction\ in\ distance\ to\ }w^*}\Big] \!=\! - \eta_t^2\cdot \underbrace{\mathbb{V}_{\mathcal{B}\sim P_t}\big[\nabla L(w_t,\mathcal{B})\big]}_{\mathrm{optimized\ through}\ P_t}\\
            & +\underbrace{2\eta_t\cdot (w_t\!-\!w^*)^\top\nabla L(w_t,\mathcal{B})\!-\eta_t^2||\nabla L(w_t,\mathcal{B})||_2^2}_{\mathrm{fixed\ by\ initial\ model\ parameter\ }w_t\mathrm{\ in\ each\ round\ t}},
        \end{aligned}
        \end{small}
    \label{eq: model convergence rate}
        \nonumber
    \end{equation}
    where $\mathbb{V}_{\mathcal{B}\sim P_t}[\nabla L(w_t,\mathcal{B})]$ denotes the gradient variance of $\mathcal{B}$.
    \begin{proof}
        The detailed proof please refer to Appendix \ref{appendix: proof of Theorem 1}.
    \end{proof}
\end{theorem}
Accordingly, {\sf IS} proposed to minimize such variance and maximize training performance by optimizing the selection probability of each data sample, i.e. $P_t(x,y)$.
However, we identify in Lemma \ref{lemma: optimal distribution for sgd} that {\sf IS} potentially assumes that each data sample in the training data batch is independently selected, resulting in optimal sample-level data selection but sub-optimal batch-level data selection for mini-batch SGD.
\begin{lemma}[Optimal Sample-Level Selection]
    \label{lemma: optimal distribution for sgd}
    To minimize the gradient variance of selected data batch $\mathcal{B}$, {\sf IS} computes the optimal selection probability $P_t^*$ for each data sample $(x,y)$:
    \begin{equation}\label{eq: data selection for sgd}
        \begin{aligned}
        P_t^*(x,y) & \triangleq 
        \mathop{\arg\min}_{P_t}\mathbb{V}_{\mathcal{B}\sim P_t}\big[\nabla L(w_t,\mathcal{B})\big]\\
        & \overset{(a)}{=} \mathop{\arg\min}_{P_t}\frac{1}{|\mathcal{B}|}\mathbb{V}_{(x,y)\sim 
        P_t}\big[\nabla l(w_t,x,y)\big]\\
        & \overset{(b)}{=} 
        \big|\big|\nabla l(w_t,x,y)\big|\big|_2\ /\sum_{(x',y')\in\mathcal{S}}\big|\big|\nabla l(w_t,x',y')\big|\big|_2, 
        \end{aligned}
        \nonumber
    \end{equation}
    where we observe that Equation (a) \textit{implicitly assumes an independent selection process for each data sample $(x,y)$ in data batch $\mathcal{B}$}. Equation (b) holds according to Cauchy-Schwarz inequality~\cite{DBLP:conf/icml/KatharopoulosF18, DBLP:conf/icml/ZhaoZ15}.
\end{lemma}
To analyze the sub-optimality of {\sf IS} in on-device settings and provide theoretical insight for designing optimal batch-level data selection strategy, we decompose the gradient variance of the selected data batch into three terms in Theorem \ref{theorem: variance decomposition for mini-batch sgd}.
\begin{theorem}[Gradient Variance Decomposition]
    The gradient variance of data batch $\mathcal{B}$ selected from candidate dataset $\mathcal{S}$ using selection strategy $P_t$ can be decomposed into the weighted sum of terms $\alpha_y,\beta_y$ and $\gamma_y$ for each class $y\!\in\!\mathcal{Y}$:
    \begin{equation}
        \begin{small}
        \begin{aligned}
            & \mathbb{V}_{\mathcal{B}\sim P_t}\big[\nabla L(w_t,\mathcal{B})\big]
            = \sum_{y\in \mathcal{Y}}\alpha_y\cdot (\beta_y\!-\!\gamma_y),
            \mathrm{where}\ 
            \alpha_y=\frac{|\mathcal{S}_y|^2}{|\mathcal{S}|^2\cdot |\mathcal{B}_y|}, \\
            & \beta_y\!=\!\sum_{(x,y)\in\mathcal{S}_y}\frac{\big|\big|\nabla l(w_t,x,y)\big|\big|^2}{|\mathcal{S}_y|^2\cdot P_{t,y}(x)},
            \gamma_y\!=\!\Big|\Big|\mathbb{E}_{(x,y)\in\mathcal{S}_y}\big[\nabla l(w_t,x,y)\big]\Big|\Big|^2,
        \end{aligned}
        \end{small}
        \label{eq: variance decomposition for mini-batch SGD}
        \nonumber
    \end{equation}
    where $\mathcal{S}_y$ and $\mathcal{B}_y$ are candidate data and selected data for each class.
    \begin{proof}
        The detailed proof please refer to Appendix \ref{appendix: proof of Theorem 2}.
    \end{proof}
    \label{theorem: variance decomposition for mini-batch sgd}
\end{theorem}

\begin{figure*}
    \subfigure[Optimality of {\sf C-IS} over random sampling ({\sf RS}) and {\sf IS}.]{
        \includegraphics[height=3cm]{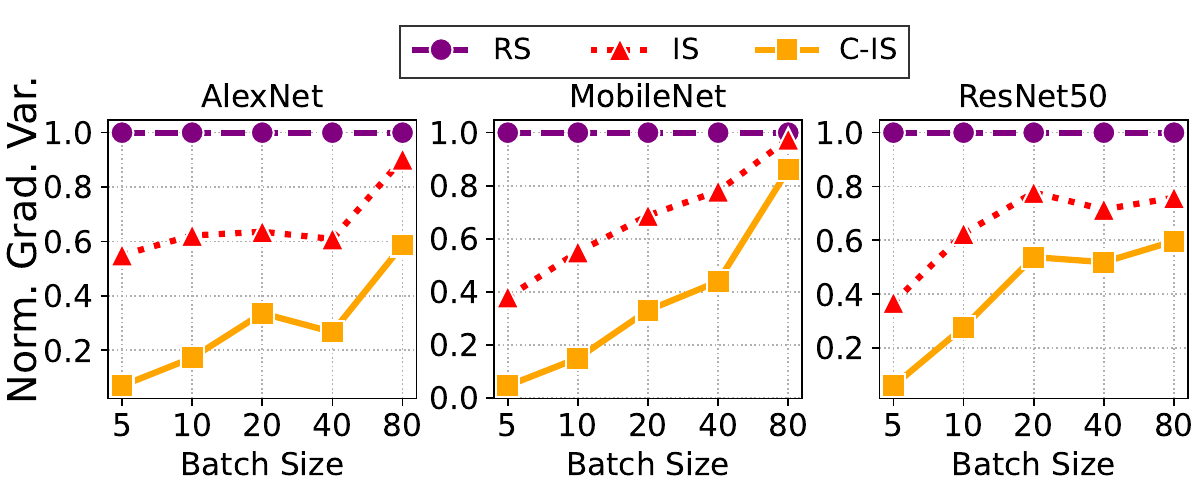}
        \label{fig: gradient variances of different methods}
    }
    \ \ 
    \subfigure[Significance of coarse-grained filter on {\sf C-IS} performance.]{
        \includegraphics[height=2.5cm]{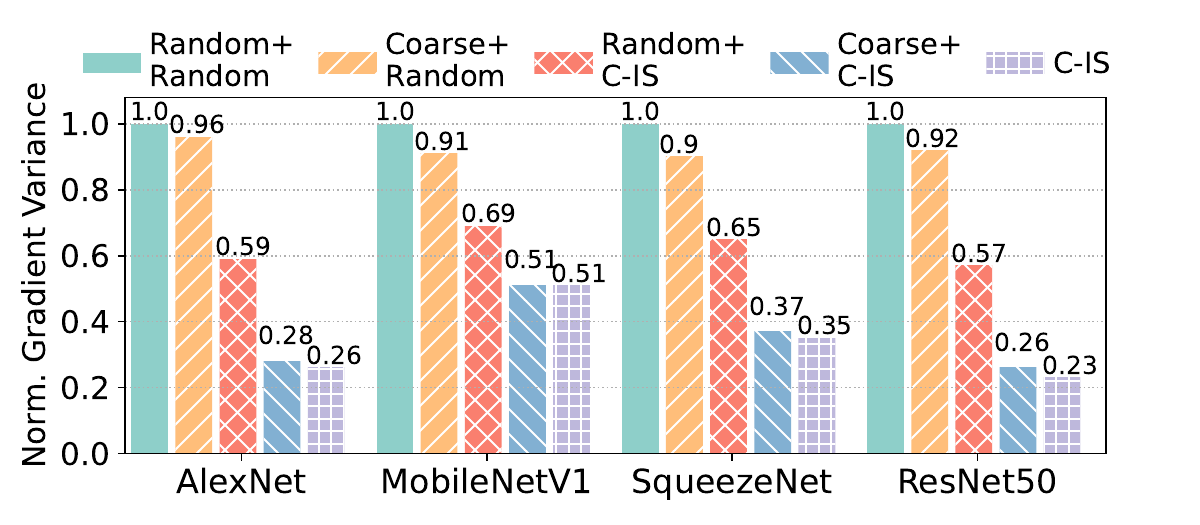}
        \label{fig: ablation study estimator error}
    }
    \ \ 
    \subfigure[Stable data importance.]{
        \includegraphics[height=2.5cm]{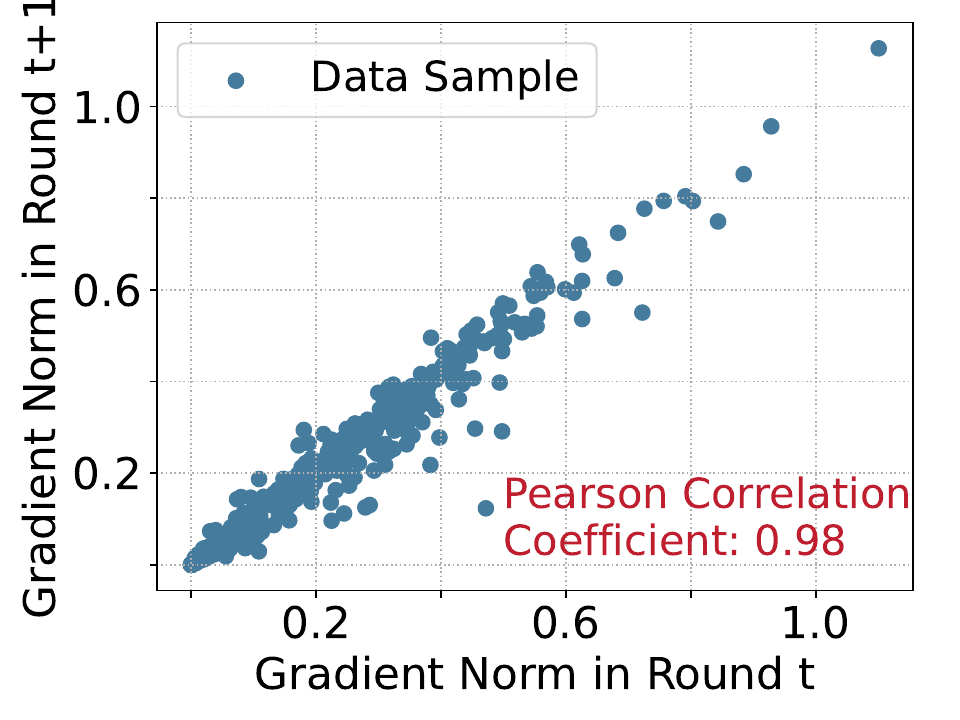}
        \label{fig: round t vs round t+1}
    }
    \vspace{-0.2cm}
    \caption{Preliminary experiments on CIFAR-10 dataset and different models with learning rate 0.1 to support some claims, including (a) to empirically demonstrate {\sf C-IS}'s optimality over previous methods, (b) to reveal the efficiency and efficacy of coarse-grained filter, and (c) to show the stable importance scores of data samples across consecutive training rounds.}
    \Description{Preliminary experiments on CIFAR-10 dataset and different models with learning rate 0.1 to support some claims, including (a) to empirically demonstrate {\sf C-IS}'s optimality over previous methods, (b) to reveal the efficiency and efficacy of coarse-grained filter, and (c) to show the stable importance scores of data samples across consecutive training rounds.}
    \vspace{-0.2cm}
\end{figure*}

\noindent We identify that the gradient variance of the selected data batch is composed of three terms of each class $y$: \textit{1)} $\alpha_y$ is impacted by batch size allocation $|\mathcal{B}_y|$ across classes and the other two terms, \textit{2)} $\beta_y$ is determined by intra-class data selection strategy $P_{t,y}$, and \textit{3)} $\gamma_y$ is a constant that varies for different classes.
As a result, traditional {\sf IS} can be regarded as conducting optimal intra-class data selection to minimize $\beta_y$, but executing sub-optimal inter-class batch size allocation based on solely $\beta_y$ rather than $(\beta_y\!-\!\gamma_y)$.
Furthermore, the overlooked term  $-\alpha_y\gamma_y$ can become significant for on-device settings with limited memory, as $\alpha_y$ increases with smaller batch sizes.
This is also verified by our empirical results in Figure \ref{fig: gradient variances of different methods}, which indicates that \textit{1)} the gradient variance gap between existing {\sf IS} and our proposed {\sf C-IS} becomes wider with smaller batch sizes and
\textit{2)} {\sf C-IS} consistently achieves the optimal performance. 

To optimize on-device model training performance, we propose a new optimal batch-level data selection strategy {\sf C-IS}, which keeps using {\sf IS} for optimal intra-class data selection while taking the integral term $(\beta_y\!-\!\gamma_y)$ into consideration when allocating batch size to different classes $y\in\mathcal{Y}$.
\begin{lemma}[Optimal Batch-Level Selection]
    To maximize the training performance of mini-batch SGD, given batch size $|\mathcal{B}|$ and dataset $\mathcal{S}_y$ for each class $y\!\in\!\mathcal{Y}$, the optimal selection size for each class (i.e. $|\mathcal{B}_y|^*$) and the optimal selection probability for each sample within the class (i.e. $P_{t,y}^*(x)$) are:
    \begin{equation}
        \begin{aligned}
            & |\mathcal{B}_y|^*\propto I_t(y), P_{t, y}^*(x)\propto I_t(x,y),
        \end{aligned}
        \nonumber
    \end{equation}
    where $I_t(y)$ and $I_t(x,y)$ are the class importance and sample importance defined in Equation (\ref{eq: class importance}) and Equation (\ref{eq: sample importance}), respectively. 
    \begin{proof}
        The detailed proof please refer to Appendix \ref{appendix: proof of Lemma 2}.
    \end{proof}
    \label{lemma: optimal distribution for minibatch sgd}
\end{lemma}

\textbf{Intuitive Understanding.} 
The sample importance $I_t(x,y)$ in Equation (\ref{eq: sample importance}) is exactly the norm of sample gradient over model parameters, reflecting the contribution of each sample to parameter update.
The class importance $I_t(y)$ in Equation (\ref{eq: class importance}) essentially quantifies the overall diversity of each class, (\textit{i.e.} gradient variance minus gradient norm variance). 
Higher class importance indicates that data samples within this class have diverse gradients but similar gradient norms. Naturally, more samples are needed to thoroughly represent the gradient distribution of such class. 
However, conventional {\sf IS} distributes batch size to each class solely based on average gradient norm, focusing on classes with high gradient value rather than diversity. 
A simple example is provided in Figure \ref{fig: simple example} for better comparison, where {\sf IS} will select the same number of samples from classes $1$ and $2$ but {\sf C-IS} will select more samples from class $1$ by considering variance, which is obviously more reasonable.

\textbf{Practical Implementation.}
For on-device implementation, we propose to substitute the gradient of each data sample over entire model parameters with the gradient over only last model layer, which avoids the cumbersome backpropagation process and saves computation and memory costs.
Such simplification relies on the phenomenon that partial model gradient can reflect the trend of full model gradient, as analyzed theoretically and empirically by existing works~\cite{DBLP:conf/icml/MirzasoleimanBL20, DBLP:conf/sigmod/LiSC22, DBLP:journals/corr/KatharopoulosF17}.

\subsection{Coarse-Grained Data Filter}
\label{sec: coarse-grained data selection}
While {\sf C-IS} enables identifying the data batch with the highest effectiveness on model performance improvement, it also incurs substantial delay due to calculating the accurate data importance (\textit{i.e.} gradient and its norm) for each streaming data. 
A straightforward remedy is to reduce the frequency of importance computation, which in turn constrains the size of candidate data for {\sf C-IS} and compromises its effectiveness. 

Inspired by the billion-scale item ranking process of online recommendation system~\cite{eksombatchai2018pixie}, {\sf Titan} leverages a two-stage architecture to guarantee both efficiency and effectiveness.
Specifically, {\sf Titan} introduces an additional stage of coarse-grained filter and designs two heuristic metrics to filter out a candidate dataset that could facilitate the processes of inter-class batch size allocation and intra-class data selection in the subsequent fine-grained selection.

\textit{1) Representativeness}: 
    To enable an accurate measurement of \textit{class importance} during inter-class batch size allocation, the filtered data is expected to \textit{represent} the characteristics of the majority of data samples in each class. 
    Therefore, the representativeness of each data$(x,y)$ can be measured by its closeness to the class centroid in feature space:
    \begin{equation}
            \mathrm{Rep}(x, y)=-\Big|\Big| f_{w}(x)-\mathbb{E}_{\mathcal{P}(x'|y)}\big[f_{w}(x')\big]\Big|\Big|_2^2,
            \label{def: representativeness}\nonumber
    \end{equation}
    where $f_w(x)$ denotes the feature extracted by current model $w$ and $\mathbb{E}_{\mathcal{P}(x'|y)}\big[f_{w}(x')\big]$ denotes the feature centroid of data in class $y$.
    
\textit{2) Diversity}: 
    To identify more high-importance data samples, the filtered data needs to be \textit{diverse} enough to cover the data distribution.
    Therefore, the \textit{diversity} of each data $(x,y)$ can be quantified as its average distance to the other data within the same class in the feature space:
    \begin{equation}
        \begin{small}
        \begin{aligned}
            & \mathrm{Div}(x,y)=\mathbb{E}_{\mathcal{P}(x'|y)}\Big[\big|\big| f_w(x)-f_w(x')\big|\big|^2_2\Big]\\
            &=\!\big|\big|f_{w}(x)\big|\big|_2^2\!+\!\mathbb{E}_{\mathcal{P}(x'|y)}\big|\big| f_w(x')\big|\big|^2_2 \!-\! 2\Big\langle f_w(x), \mathbb{E}_{\mathcal{P}(x'|y)}\big[f_w(x')\big]\Big\rangle.\nonumber
        \end{aligned}
        \end{small}
        \label{def: diversity}
    \end{equation}

We evaluate the impact of coarse-grained filter on {\sf C-IS}'s ability in reducing gradient variance in Figure \ref{fig: ablation study estimator error}, where $A\!+\!B$ denotes leveraging method $A$ to filter out $0.3v$ candidate data samples out of $v$ streaming samples and employing method $B$ to further select $0.1v$ data samples as data batch, where $v$ is set to 100 for preliminary experiments.
The result shows that compared with the ideal case of performing {\sf C-IS} on all data, coarse-grained filter can reduce the candidate data size by $70\%$ with less than $3\%$ degradation of gradient variance reduction degree.

\textbf{Practical Implementation.}
For efficient implementation of coarse-grained filter, {\sf Titan} only needs to dynamically maintain two running-sum estimators for average feature {\small $\mathbb{E}_{\mathcal{P}(x'|y)}\big[f_w(x')\big]$} and average feature norm {\small $\mathbb{E}_{\mathcal{P}(x'|y)}\big|\big| f_w(x')\big|\big|_2^2$} using each streaming data $(x,y)$.
Based on these estimators, {\sf Titan} could realize online coarse-grained filtering by buffering data with the highest $\mathrm{Rep}(x,y)\!+\!\mathrm{Div}(x,y)$.
For the feature extraction function $f_w(x)$, we propose to input the raw data $x$ into the first few network layers of model $w$ and regard the layer outputs as features, which is according to our empirical observation that 
\textit{1)} the features extracted by shallow layers are sufficient to filter out an effective candidate data for subsequent fine-grained data selection, and 
\textit{2)} forward pass through a few model layers only introduces minor latency and memory footprint. A detailed empirical analysis is presented in \S\ref{sec: Component-Wise Analysis}.

\subsection{Pipeline Design} 
\label{sec: pipeline design}
Although the two-stage architecture of {\sf Titan} achieves higher time-efficiency, the wall-clock time per training round still increases significantly due to the model dependency and resource preemption between data selection and model update:
\textit{1) Model dependency}: As data selection relies on the latest model parameter to compute the accurate importance of each sample and class, the processes of model update and data selection have to be executed alternately and sequentially.
\textit{2) Resource preemption}: The limited computing resource is shared and preempted by data selection and model update, which will slow down the original model update process.

{\sffamily Titan} overcomes the above challenges by leveraging a pipeline design to enable the co-execution of model update and data selection.
To eliminate \textit{model dependency}, {\sf Titan} proposes a simple but effective ``one-round-delay'' scheme, where each model parameter $w_t$ is updated by the data batch selected in the previous round using the slightly outdated model $w_{t-1}$. Such approximation enables the co-execution of parameter update for the current round and data selection for the next round, and its feasibility is supported by our observation in Figure \ref{fig: round t vs round t+1} that per-sample importance (\textit{i.e.} gradient norm) typically does not change significantly in consecutive training rounds.

To avoid \textit{resource preemption}, {\sf Titan} offloads the data selection process to commonly seen idle computing resources. 
Despite that mobile devices are typically equipped with multiple types of computing resources (\textit{e.g.} CPU, GPU and NPU), current devices mainly use one type of resource type for parameter update, due to the high synchronization overhead of sharing each layer's outputs and gradients across different hardware per each parameter update~\cite{DBLP:conf/mobisys/JiaZCJLRZ22, wei2023nn}. 
By offloading only data selection to other available computing resource, {\sf Titan} prevents its resource conflict with model update and incurs low cost by synchronizing model parameters and the small selected data batch only once per model update. A breakdown analysis of the system cost is provided in Figure \ref{fig: system overhead} in \S\ref{sec: Component-Wise Analysis}.

Further, to handle dynamic idle computing resources on edge devices, {\sf Titan} system automatically adjusts the candidate data size chosen by coarse-grained filter. Specifically, it continuously evaluates the importance of each stored data sample using idle resources. For each new training round, these evaluated samples naturally become candidate data for fine-grained selection, without the need to predefine a candidate data size. Also, our evaluation already accounts for such resource fluctuations, as the computation speed is impacted by dynamic factors like device power, heat dissipation and charging.

%% file: Contents/4-Evaluation.tex
\section{Evaluation}
\label{section: evaluation}
We first introduce our experiment setup (\S\ref{sec: experiment setup}). Then we present the overall performance of {\sf Titan} (\S\ref{sec: overall performance}) and conduct component-wise analysis (\S\ref{sec: Component-Wise Analysis}).
Further, we test the applicability of {\sf Titan} to different scenarios in Appendix \ref{appendix: extended experiments}, including fluctuant idle computing resources, federated learning and noisy on-device data.

\subsection{Experiment Setup}
\label{sec: experiment setup}

\textbf{Tasks, Datasets and Models.}
To demonstrate {\sf Titan}'s generality, we evaluate it on three typical mobile computing tasks with three data modalities and six model structures:
\textit{1) Image Classification (IC)}: CIFAR-10~\cite{krizhevsky2009learning} consists of $60, 000$ images of 10 objects. We train four representative ML models for this task, including the classic dense model AlexNet~\cite{NIPS2012_c399862d}, lightweight models MobileNetV1~\cite{howard2017mobilenets} and SqueezeNet~\cite{iandola2016squeezenet} as well as a larger model ResNet50~\cite{he2016deep}.
\textit{2) Audio Recognition (AR)}: Google Speech Commands~\cite{warden2018speech} includes $100,000$ sound files of $20$ commands collected from $2,000$ users, and we train ResNet34~\cite{he2016deep} for this task. 
\textit{3) Human Activity Recognition (HAR)}: HARBOX~\cite{ouyang2021clusterfl} contains IMU data collected from 6 activities of 121 users. According to previous work, we resample with a sliding time window of $2$s at $50$Hz and obtain $34,115$ data samples with 900-dimension features. An MLP~\cite{pinkus1999approximation} with two fully-connected layers and a SoftMax layer is trained for this task.

\textbf{Hardware Setup.}
We implement {\sf Titan} framework on the real-world mobile platform NVIDIA Jetson Nano~\cite{nano} with 4GB RAM, 4 A57 CPU cores and a Maxwell GPU, which has similar hardware and running environment with mainstream devices~\cite{DBLP:conf/mobicom/LiZZC22, cold_start}. 
For pipeline implementation, {\sf Titan} forks three processes using different computation hardware\footnote{
We use mobile CPU for model update and GPU for data selection as
\textit{1)} CPUs train models faster than GPUs on current mobile devices~\cite{DBLP:conf/mobisys/JiaZCJLRZ22, DBLP:conf/mobicom/WangDCLX21, das2022enabling}, 
\textit{2)} CPUs are more supported by today's on-device training libraries~\cite{DBLP:conf/mobisys/WangXJDY0HLL22}, and 
\textit{3)} using GPU for model update and CPU for data selection can be viewed as cases with varied amounts of idle computing resources, analyzed in Appendix \ref{appendix: extended experiments}.}: 
\textit{Process 1} conducts coarse-grained filtering with mobile GPU to filter out a small candidate dataset
from data streams;
\textit{Process 2} executes fine-grained selection with mobile GPU to identify the optimal data batch from the candidate data; 
\textit{Process 3} steadily updates the model parameter with mobile CPU using the data batch shared by process 2. 

\textbf{Baselines.}
We compare {\sf Titan} with existing data selection methods, including:
\textit{1) Random selection} ({\sf RS}) selects random data for model training;
\textit{2) Importance sampling} ({\sf IS})~\cite{DBLP:conf/icml/KatharopoulosF18} selects each training data sample according to gradient norm over the final model layer;
\textit{3) Heuristic data selection} selects training data batch according to per-sample training loss (high loss {\sf HL}~\cite{DBLP:conf/iclr/ColemanYMMBLLZ20} and low loss {\sf LL}~\cite{DBLP:conf/aistats/ShahWS20}), cross entropy of the model output logits ({\sf CE}~\cite{settles2009active}) or data representativeness and diversity ({\sf OCS}~\cite{DBLP:conf/iclr/YoonMYH22});
\textit{4) Coreset selection} ({\sf Camel}~\cite{DBLP:conf/sigmod/LiSC22}) greedily selects the sample that minimizes the input distance between the currently selected data batch and entire dataset. 

\textbf{Evaluation Metrics.}
We use five metrics to evaluate the overall performance of data selection.
\textit{1) Final inference accuracy} denotes the test accuracy of the finally trained model.
\textit{2) Time-to-accuracy} measures the wall-clock time required for each method to reach the target accuracy. For simplicity, the target accuracy is set as the final accuracy of {\sf RS}.
\textit{3) Processing latency} quantifies the time cost for processing each streaming data.
\textit{4) Memory and 5) energy consumption} measure the peak memory footprint and overall energy cost of {\sf Titan} framework for evaluating system overheads.

\textbf{Parameter Configuration.}
The default learning rates are 0.1 for AlexNet, MobileNet and SqueezeNet and 0.005 for other larger models, reduced by a factor of $0.95$ per 100 training rounds. The training batch size is 10 to satisfy the memory constraint of common devices as elaborated in \S\ref{sec: Resource Limitation Enforces On-Device Data Selection}. 
The velocity of on-device data stream is set to 100 samples per training round, indicating that 10 out of 100 streaming samples are selected as training data batch for model update in each round. 
For coarse-grained filter, we use the first model block\footnote{Current ML models typically consists of several blocks with similar structures and each block is composed of several neural network layers.} for feature extraction, and set the size budget for the buffered candidate dataset to $30$ samples.
\begin{table*}
    \centering
        \caption{Overall performance of {\sf Titan} and baselines, where \colorbox{blue!20}{blue} highlights the top value. For baselines failing to reach target accuracy, we simply present the normalized time of entire model training process.}
        \vspace{-0.1cm}
        \begin{tabular}{|c|c|p{0.5cm}<{\centering}p{0.5cm}<{\centering}p{0.5cm}<{\centering}p{0.5cm}<{\centering}p{0.5cm}<{\centering}p{0.5cm}<{\centering}p{0.6cm}<{\centering}p{0.6cm}<{\centering}|p{0.5cm}<{\centering}p{0.5cm}<{\centering}p{0.5cm}<{\centering}p{0.5cm}<{\centering}p{0.5cm}<{\centering}p{0.5cm}<{\centering}p{0.6cm}<{\centering}p{0.6cm}<{\centering}|}
            \hline
            \multirow{2}{*}{\textbf{Task}} & \multirow{2}{*}{\textbf{Model}} &
            \multicolumn{8}{c|}{\textbf{Normalized Time-to-Accuracy ($\times$)}} & \multicolumn{8}{c|}{\textbf{Final Model Accuracy ($\%$)}}\\
            & & {\sf RS} & {\sf IS} & {\sf LL} & {\sf HL} & {\sf CE} & {\sf OCS} & {\sf Camel} & {\sf Titan} & {\sf RS} & {\sf IS} & {\sf LL} & {\sf HL} & {\sf CE} & {\sf OCS} & {\sf Camel} & {\sf Titan}\\
            \hline
            \multirow{4}{*}{IC} & AlexNet & 1.00 & 3.25 & 3.98 & 3.98 & 3.59 & 4.06 & 2.07 & \cellcolor{blue!20}0.70 & 71.2 & 73.5 & 18.2 & 34.3 & 71.6 & 62.3 & 71.3 & \cellcolor{blue!20}74.5 \\
            & MobileNet & 1.00 & 3.22 & 3.45 & 3.45 & 3.41 & 3.67 & 1.15 & \cellcolor{blue!20}0.57  & 69.2 & 69.5 & 17.7 & 13.9 & 69.6 & 38.1 & 68.7 & \cellcolor{blue!20}75.4 \\
            & SqueezeNet& 1.00 & 3.96 & 3.97 & 3.97 &  3.04 & 4.06 & 2.07 & \cellcolor{blue!20}0.69  & 76.2 & 73.0 & 18.3 & 45.0 & 78.0 & 40.7 & 75.6 & \cellcolor{blue!20}79.0 \\
            & ResNet50 & 1.00 & 2.32 & 3.14 & 3.14 & 2.20 & 2.18 & 1.11 & \cellcolor{blue!20}0.66& 76.5 & 78.0 & 22.3 & 34.9 & \cellcolor{blue!20}81.7 & 27.3 & 76.8 & 81.1 \\
            \hline
            AR & ResNet34& 1.00 & 2.04 & 3.14 & 3.14 & 2.96 & 3.19 & 0.81 & \cellcolor{blue!20}0.77 & 76.0 & 78.7 & 14.7 & 58.8 & 73.2 & 59.4 & 76.5 & \cellcolor{blue!20}79.8 \\
            \hline
            HAR & MLP& 1.00 & 3.56 & 6.30 & 6.47 & 5.28 & 14.4 & 12.5 & \cellcolor{blue!20}0.71 & 75.5 & \cellcolor{blue!20}77.5 & 45.5 & 21.8 & 60.9 & 68.0 & 75.6 & 76.7 \\
            \hline
        \end{tabular}
        \label{tab: overall performance}
        \vspace{-0.2cm}
\end{table*}
\begin{figure*}
    \centering
    \subfigure[Per-round training time.]{
        \includegraphics[height=2.75cm]{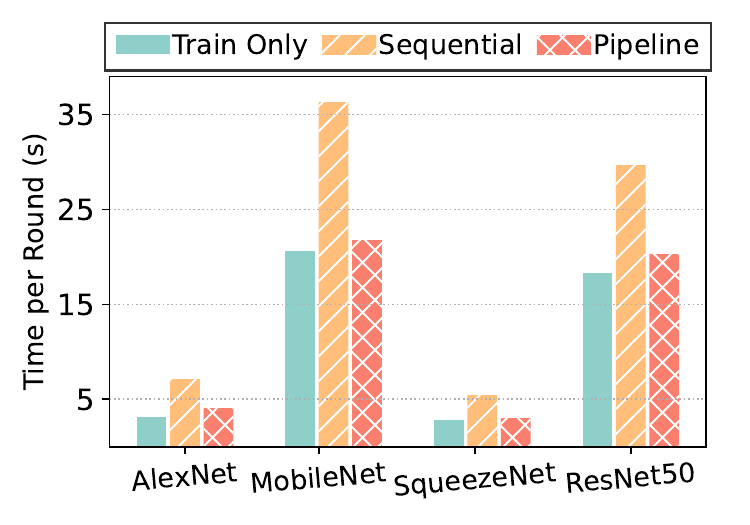}
        \label{fig: pipeline breakdown}
    }
    \subfigure[Processing delay per streaming data sample.]{
        \includegraphics[height=2.75cm]{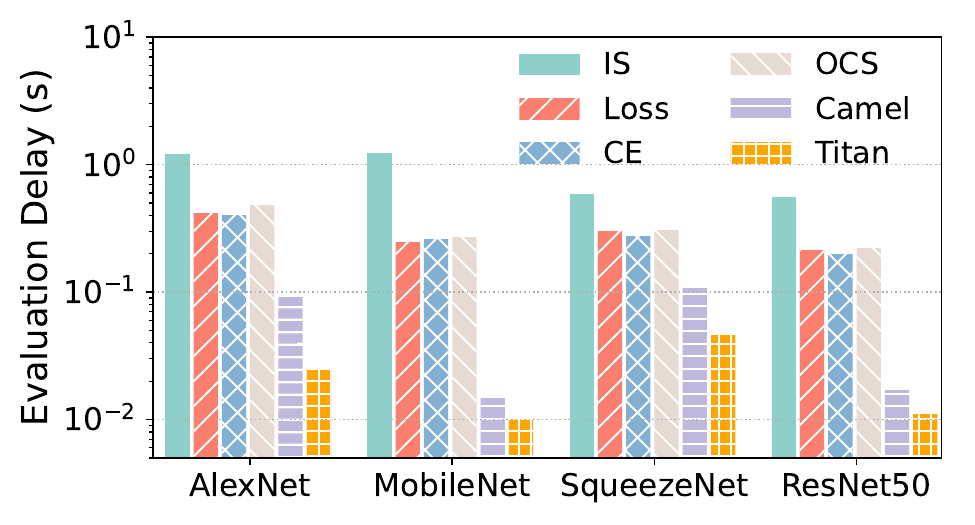}
        \label{fig: relative per-sample delay}
    }
    \subfigure[Peak memory footprint.]{
        \includegraphics[height=2.75cm]{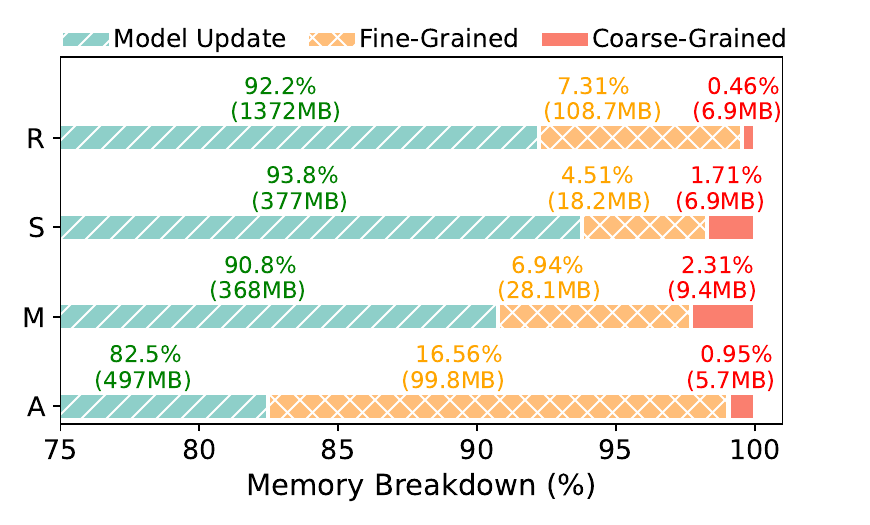}
        \label{fig: memory footprint of different methods}
    }
    \subfigure[Average energy consumption.]{
        \includegraphics[height=2.75cm]{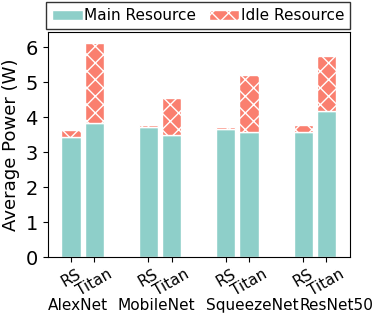}
        \label{fig: power}
    }
    \vspace{-0.3cm}
    \caption{Overall analysis of {\sf Titan}'s system overhead, including time, memory and energy.}
    \Description{Overall analysis of {\sf Titan}'s system overhead, including time, memory and energy.}
    \label{fig: system overhead}
\end{figure*}

\subsection{Overall Performance}
\label{sec: overall performance}
We begin by comparing the overall performance of {\sf Titan} with all baselines on three representative mobile computing tasks. 

\textbf{{\sf Titan} significantly reduces the wall-clock time to reach target accuracy.} 
Table \ref{tab: overall performance} summarizes the time taken by different methods to reach target accuracy, which is normalized by the time of {\sf RS} for clearer comparison.
Compared with the most lightweight baseline, {\sf Titan} reduces the training time by $30\!-\!43\%$ for IC task, $23\%$ for AR task, and $29\%$ for HAR task. 
We observe that most baselines have significantly longer training time than {\sf Titan}, caused by the extra delay of computing each streaming sample's importance and inferior improvement in model training performance. 
In contrast, {\sf Titan}'s data selection method is theoretically guaranteed to optimize the model performance in each round and its pipeline design further overlaps the extra time cost.

\textbf{{\sf Titan} maintains or improves the final inference accuracy of on-device models.} 
Table \ref{tab: overall performance} shows that {\sf Titan} achieves the highest final accuracy for most ML models, including AlexNet, MobileNetV1, SqueezeNet and ResNet34, and achieves the second-best accuracy on other ML models with only marginal accuracy drop compared to the top baseline, such as $0.6\%$ drop compared to {\sf CE} on ResNet50 and $0.8\%$ drop compared to {\sf IS} on MLP.

\textbf{{\sf Titan} reduces the processing time of each streaming data to millisecond-level.} 
As shown in Figure \ref{fig: relative per-sample delay}, {\sf Titan} achieves the lowest processing time and highest throughput for data importance computation, with only $4\!-\!13$ms across different model structures.
The millisecond-level latency is attributed to the time-efficiency of coarse-grained filter, and further facilitates the practical deployment of {\sf Titan} in common applications without compromising the quality of real-time service.

\textbf{{\sf Titan} introduces marginal system overheads, including peak memory footprint and overall energy consumption.} 
Figure \ref{fig: memory footprint of different methods} breaks down the memory footprint of {\sf Titan}, indicating that the pipeline design incurs less than $10\%$ extra memory footprint compared with original model training process, such as $105$MB, $37$MB, $25$MB and $114$MB for AlexNet, MobileNet,  SqueezeNet and ResNet50. 
The high memory costs for AlexNet and ResNet50 are attributed to their large parameter sizes, and for lightweight models like MobileNet and SqueezeNet, {\sf Titan} incurs less than $40$MB memory overhead.
Figure \ref{fig: power} compares the average device power of original model training (\textit{i.e.} {\sf RS}) and {\sf Titan} framework. We notice that while our system increases device power (1.68$\times$, 1.21$\times$, 1.39$\times$, and 1.53$\times$ higher than RS for four models), it reduces the wall-clock training time (0.7$\times$, 0.57$\times$, 0.69$\times$, and 0.66$\times$). As a result, the total energy consumption becomes 1.17$\times$, 0.69$\times$, 0.96$\times$, and 1.01$\times$ compared to RS, reflecting a 31\% reduction to 18\% increase.
Accordingly, {\sf Titan} leads to only marginal or even reduced energy costs, but achieves much less training time and higher model accuracy.

\subsection{Component-Wise Analysis}
\label{sec: Component-Wise Analysis}
We then analyze the role of each key component in {\sf Titan}, including fine-grained selection, coarse-grained filter and pipeline design.

\begin{figure}
    \centering
    \includegraphics[width=0.4\linewidth]{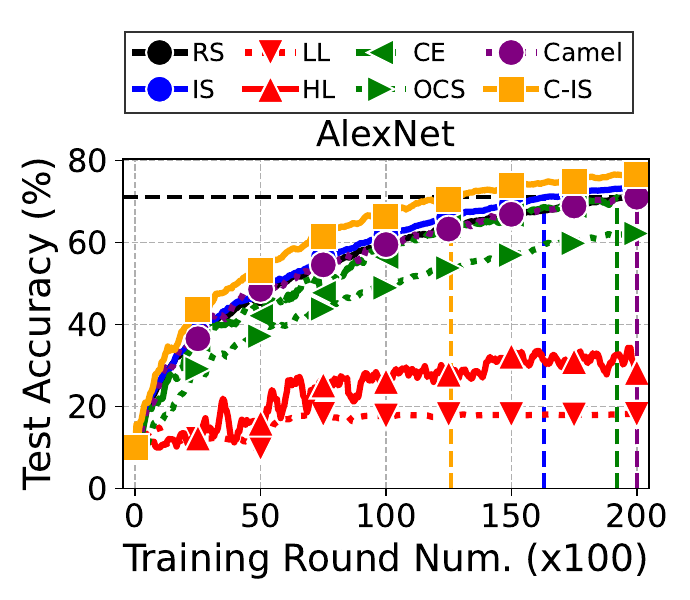}
    \ \ \ \ 
    \includegraphics[width=0.4\linewidth]{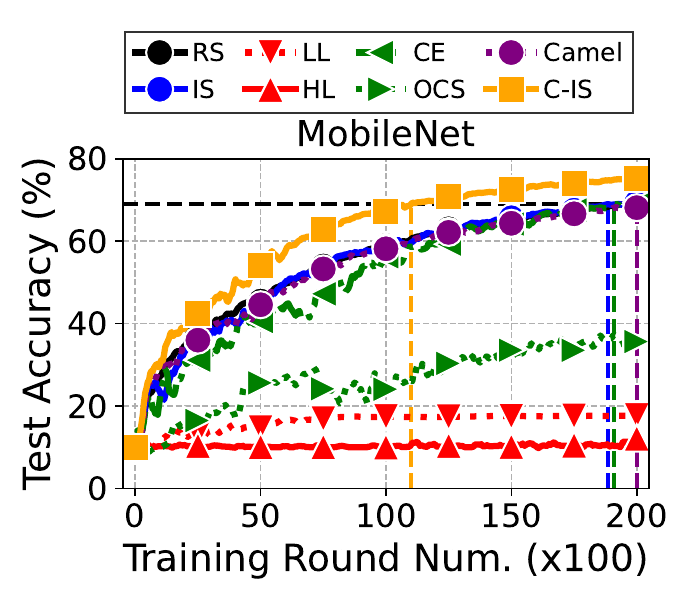}
    \includegraphics[width=0.4\linewidth]{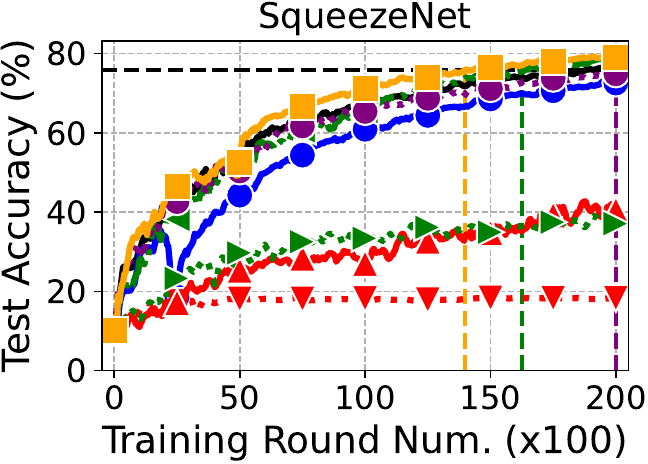}
    \ \ \ \ 
    \includegraphics[width=0.4\linewidth]{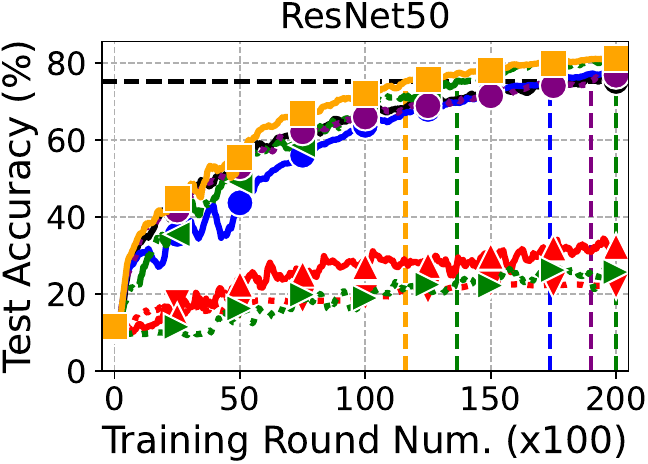}
    \vspace{-0.3cm}
    \caption{Training curves of different methods. Horizontal line denotes target accuracy, and vertical lines denote numbers of rounds to reach such accuracy.}
    \Description{Training curves of different methods. Horizontal line denotes target accuracy, and vertical lines denote numbers of rounds to reach such accuracy.}
    \vspace{-0.2cm}
    \label{fig: effect of fine-grained data selection}
\end{figure}

\textbf{Fine-Grained Selection.}
To show the individual impact of fine-grained selection strategy {\sf C-IS}, we compare the training processes of different data selection methods in Figure \ref{fig: effect of fine-grained data selection}. 
Across various model structures, {\sf C-IS} consistently achieves the best model training performance, with $5.8\%$ increase in final accuracy and $1.59\times$ speedup in model convergence rate on AlexNet, $4.8\%$ and $1.62\times$ on MobileNetV1, $3.1\%$ and $1.43\times$ on SqueezeNet, $4.9\%$ and $1.72\times$ on ResNet50, which coincides the theoretical optimality of {\sf C-IS} analyzed in \S\ref{sec: fine-grained data selection}.

\begin{figure}
    \centering
    \includegraphics[width=\linewidth]{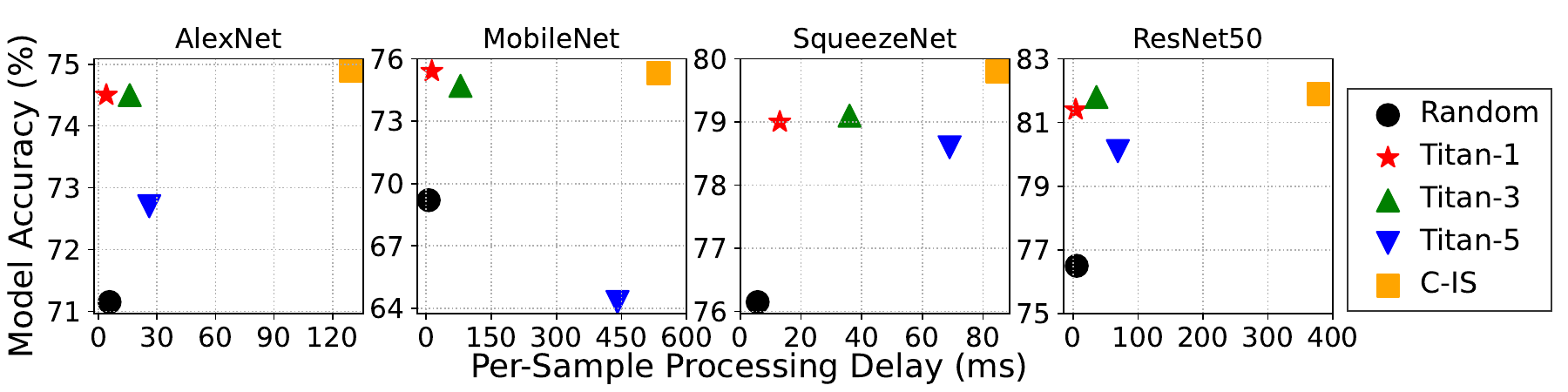}
    \vspace{-0.4cm}
    \caption{Impact of coarse-grained filter on data processing delay and final model accuracy across different models and model block numbers for feature extraction.}
    \Description{Impact of coarse-grained filter on data processing delay and final model accuracy across different models and model block numbers for feature extraction.}
    \vspace{-0.3cm}
    \label{fig: two-stage data selection}
\end{figure}

\textbf{Coarse-Grained Filter.}
We further conduct a comparison between only individual fine-grained selection ({\sf C-IS}) and {\sf Titan} with different number $n$ of model blocks for feature extraction in coarse-grained filter ({\sf Titan}-$n$).
Empirical results in Figure \ref{fig: two-stage data selection} reveal the following results:
\textit{1)} Compared with {\sf C-IS}, coarse-grained filter significantly reduces the processing delay of each streaming data, achieving speedup of $32\times$, $40\times$, $6.5\times$, $94\times$ on AlexNet, MobileNetV1, SqueezeNet and ResNet50;
\textit{2)} The shallow features extracted by the first model block exhibit satisfactory performance in selecting a candidate dataset with potential high importance for fine-grained data selection, with only $0.1\%\!-\!0.4\%$ model accuracy drop in various model structures, compared with the ideal case of conducting {\sf C-IS} on all streaming data;
\textit{3)} When leveraging more model blocks for feature extraction, the effectiveness of {\sf Titan} seems to gradually degrade. This is because deeper model layers tend to extract more concentrated and similar features for data samples within the same class, making it more difficult to filter out diverse data for intra-class data selection.
Consequently, we propose to leverage the first model block in practice for high time-efficiency and stable training performance improvement.

\textbf{Pipeline Design.}
To show the role of the pipeline design in reducing time overhead and resource conflict, we visualize the per-round time of only model training, sequential execution and co-execution of model update and data selection in Figure \ref{fig: pipeline breakdown}, which demonstrates that pipeline incurs negligible time for synchronizing the model and data between different processes compared with practical model training time.

\subsection{Application to Various Scenarios.}
In Appendix \ref{appendix: extended experiments}, we provide the empirical results when applying {\sf Titan} to various practical scenarios, including fluctuant idle computing resources, federated learning and noisy on-device data streams.

%% file: Contents/5-RelatedWork.tex
\section{Other Related Work}
In \S\ref{sec: limitation of state-of-the-art}, we provided a thorough introduction of other data selection approaches and here we review other relevant works.

\textbf{On-Device Model Training.}
Recently, there has been a trend towards moving model training from cloud servers to the resource-constrained devices. Prior works focused on improving the utilization of hardware resource (\textit{e.g.}, memory, storage and computing resource) to enhance training efficiency, such as optimizing memory allocation to increase training batch size~\cite{DBLP:conf/mobisys/WangXJDY0HLL22, DBLP:conf/mobisys/GimK22}, exploring the co-execution of multiple types of computing resources to accelerate computation~\cite{DBLP:conf/mobicom/XuXWW0H0JL22, DBLP:conf/mobisys/JiaZCJLRZ22, DBLP:conf/mobicom/WangDCLX21}, and offloading computation to cloud server~\cite{yao2021context,wang2020surveiledge,pang2022towards}. 
However, only a few works noticed the under-utilization of on-device data resource, which address such issue through cloud-side data distribution estimation and model pre-training~\cite{DBLP:journals/imwut/XuQMHL18, liu2020pmc} before model deployment.
Therefore, {\sf Titan} is complementary to previous works.

\textbf{Two-Stage Architecture.} The design of two-stage system has been widely adopted in industrial recommendation system~\cite{eksombatchai2018pixie, DBLP:conf/recsys/CovingtonAS16, DBLP:conf/kdd/BorisyukKSZ16, DBLP:conf/www/EksombatchaiJLL18} to recommend highly personalized items from a vast item space in real-time. In the first stage, one or multiple efficient retrieval models are used to produce a candidate set that contains thousands of items from the whole item space. Then, in the second stage, a more powerful model re-ranks the candidate items and recommends the top few items to the user. 
Such design allows for a trade-off between the system scalability and performance. 
In the area of on-device data selection, to the best of our knowledge, we are the first to consider leveraging the two-stage design to simultaneously enhance the effectiveness and efficiency of data utilization to improve model training performance.

\textbf{Difference with Federated Learning (FL).}
For motivation and setting, we target model training on a single device using local data, while FL trains model across multiple devices with heterogeneous data distributions. As a result, FL focuses on mitigating the negative impact of cross-device data heterogeneity on global model performance, and does not fully consider the data utilization for local model training on each single device. In contrast, we tackle data under-utilization problem in a broader scenario of on-device model training with both theoretical and empirical analysis, which has not been studied before to the best of our knowledge. In Appendix \ref{appendix: extended experiments}, we extend our evaluation to an FL setting with 50 devices and heterogeneous local data distributions, where we achieve a 3.17$\times$ speedup in global model convergence and a 2.03\% increase in accuracy. However, FL-specific data utilization methods are not applicable to our single-device setting, as they aim to tackle non-i.i.d data distribution across devices, which rely on the presence of multiple devices.

%% file: Contents/6-Conclusion.tex
\section{Conclusion}
\label{section: conclusion}
In this work, we identify that the under-utilization of on-device data resource hinders the successful model training process for edge computing tasks. To address this issue, we propose an on-device data selection framework {\sf Titan} to simultaneously achieve high effectiveness and time-and-resource efficiency for on-device data utilization through an optimal data selection strategy, a two-stage architecture and a pipeline design. Extensive evaluation on real-world device and representative edge computing tasks demonstrate the remarkable advantages of {\sf Titan} in final model accuracy and wall-clock training time compared with conventional cloud-side data selection approaches, with minor additional system costs.

%% file: Contents/7-Appendix.tex
\appendix

\section{Proofs}
\subsection{Proof of Theorem \ref{theorem: convergence rate}}
\label{appendix: proof of Theorem 1}
According to the model update formula in Equation (\ref{eq: model update}), we have $w_{t+1}\!=\!w_t\!-\!\eta_t\nabla L(\mathcal{B},w_t)$, which leads to:
    \begin{equation}
        \begin{small}
        \begin{aligned}
        &\mathbb{E}_{\mathcal{B}\sim P_t(\mathcal{S})}\Big[||w_{t}\!-\!w^*||_2^2-|| w_{t+1}\!-\!w^*||_2^2\Big]\\
        = & \mathbb{E}_{\mathcal{B}\sim P_t(\mathcal{S})}[w_{t}^\top w_{t}-2w_{t}^\top w^*+2w_{t+1}^\top w^*-w_{t+1}^\top w_{t+1}]\\
        = & \mathbb{E}_{\mathcal{B}\sim P_t(\mathcal{S})}\Big[w_{t}^\top w_{t}\!-\!2w_{t}^\top w^*\!+2\big(w_t\!-\!\eta_t\nabla L(\mathcal{B},w_t)\big)^\top w^*\\
        & -\!\big(w_t\!-\!\eta_t\nabla L(\mathcal{B},w_t)\big)^\top\big(w_t\!-\!\eta_t\nabla L(\mathcal{B},w_t)\big)\Big]\\
        = & \mathbb{E}_{\mathcal{B}\sim P_t(\mathcal{S})}\Big[-2\eta_t\cdot\nabla L(\mathcal{B},w_t)^\top(w^*\!-\!w_t)-\eta_t^2\big|\big|\nabla L(\mathcal{B},w_t
        )\big|\big|^2\Big]\\
        \overset{(a)}{=} & -2\eta_t\cdot(w^*\!-\!w_t)^\top\nabla L(\mathcal{S},w_t)+\eta_t^2\cdot\mathbb{E}_{\mathcal{B}\sim P_t(\mathcal{S})}\big|\big|\nabla L(\mathcal{B},w_t)\big|\big|^2,
        \end{aligned}
        \end{small}
        \nonumber
    \end{equation}
    where equality (a) holds because of the unbiased data sampling $\mathcal{B}\sim P_t(\mathcal{S})$. 
    Next, we leverage the definition of variance, i.e., $\mathbb{V}[x]=\mathbb{E}[x^2]-\big(\mathbb{E}[x]\big)^2$ to further decompose the final term:
    \begin{equation}
        \begin{small}
        \begin{aligned}
            & \mathbb{E}_{\mathcal{B}\sim P_t(\mathcal{S})}\big|\big|\nabla L(\mathcal{B},w_t)\big|\big|^2\\
            = & \mathbb{V}_{\mathcal{B}\sim P_t(\mathcal{S})}\big[\nabla L(\mathcal{B},w_t)\big] + \Big|\Big|\mathbb{E}_{\mathcal{B}\sim P_t(\mathcal{S})}\big[\nabla L(\mathcal{B},w_t)\big]\Big|\Big|^2\\
            = & \mathbb{V}_{\mathcal{B}\sim P_t(\mathcal{S})}\big[\nabla L(\mathcal{B},w_t)\big] + \big|\big|\nabla L(\mathcal{S},w_t)\big|\big|^2,
        \end{aligned}
        \end{small}
        \nonumber
    \end{equation}
    which leads to the conclusion in Theorem \ref{theorem: convergence rate}.    

\subsection{Proof of Theorem \ref{theorem: variance decomposition for mini-batch sgd}}
\label{appendix: proof of Theorem 2}
We first decompose the gradient variance of the data batch $\mathcal{B}$ selected from dataset $\mathcal{S}$ into the weighted variances of sub-batch $\mathcal{B}_y$ selected from data-subset $\mathcal{S}_y$ for each class $y\in\mathcal{Y}$:
        \begin{equation}
            \begin{small}
            \begin{aligned}
                & \mathbb{V}_{\mathcal{B}\sim P_t(\mathcal{S})}\!\left[\nabla\! L(w,\mathcal{B})\right]
                \!=\! \mathbb{V}_{\mathcal{B}\sim P_t(\mathcal{S})}\!\!\bigg[\sum_{y\in\mathcal{Y}}\frac{|\mathcal{S}_y|}{|\mathcal{S}|}\mathbb{E}_{(x,y)\in\mathcal{B}_y}\!\left[\nabla l(w,x,y)\right]\bigg]\\
                & = \sum_{y\in\mathcal{Y}}\frac{|\mathcal{S}_y|^2}{|\mathcal{S}|^2} \mathbb{V}_{\mathcal{B}_y\sim P_{t,y}(\mathcal{S}_y)}\left[\mathbb{E}_{(x,y)\in\mathcal{B}_y}\left[\nabla l(w,x,y)\right]\right]\ \ \ \ \ \ \ \ \ \ \ \ \ \ \ \mathrm{(d)}\\
                & = \sum_{y\in\mathcal{Y}}\frac{|\mathcal{S}_y|^2}{|\mathcal{S}|^2\cdot|\mathcal{B}_y|}\mathbb{V}_{(x,y)\sim P_{t,y}(\mathcal{S}_y)}\left[\nabla l(w,x,y)\right]\ \ \ \ \ \ \ \ \ \ \ \ \ \ \ \ \ \ \ \ \ \ \ \ \mathrm{(e)}.
            \end{aligned}
            \end{small}
            \nonumber
        \end{equation}
        Equation(d) decomposes the overall batch selection process into sub-processes for each class, and Equation(e) holds because each sample in sub-batch $\mathcal{B}_y$ is selected from $\mathcal{S}_y$ with strategy $P_{t,y}$.
        According to the variance definition (\textit{i.e.}$\mathbb{V}[x]=\mathbb{E}[x^2]\!-\!\big(\mathbb{E}[x]\big)^2$), we can further decompose the gradient variance:
        \begin{equation}
            \begin{small}
            \begin{aligned}
                \mathrm{(e)} = &\sum_{y\in\mathcal{Y}}\frac{|\mathcal{S}_y|^2}{|\mathcal{S}|^2\cdot| \mathcal{B}_y|}\cdot\Bigg[\sum_{(x,y)\in\mathcal{S}_y}\!P_{t,y}(x)\cdot \frac{\big|\big|\nabla l(w,x,y)\big|\big|^2}{\big[P_{t,y}(x)\cdot |\mathcal{S}_y|\big]^2}-\\
                &\bigg|\bigg|\sum_{(x,y)\in\mathcal{S}_y}P_{t,y}(x)\cdot\frac{\nabla l(w,x,y)}{P_{t,y}(x)\cdot|\mathcal{S}_y|}\bigg|\bigg|^2\Bigg]\ \ \ \ \ \ \ \ \ \ \ \ \ \ \ \ \ \ \ \ \ \ \ \ \ \quad\quad\mathrm{(f)}\\
                = & \sum_{y\in\mathcal{Y}}\frac{|\mathcal{S}_y|^2}{|\mathcal{S}|^2\cdot|\mathcal{B}_y|}\cdot \bigg[\sum_{(x,y)\in\mathcal{S}_y}\frac{\big|\big|\nabla l(w,x,t)\big|\big|^2}{|\mathcal{S}_y|^2\cdot P_{t,y}(x)}-\\
                & \big|\big|\frac{\sum_{(x,y)\in\mathcal{S}_y}\nabla l(w,x,y)}{|\mathcal{S}_y|}\big|\big|^2\bigg]
                = \sum_{y\in\mathcal{Y}}\alpha_y\cdot\big(\beta_y-\gamma_y\big).
            \end{aligned}
            \end{small}
            \nonumber
        \end{equation}
        Equation(f) holds because in data selection, to ensure the unbiasedness of selected data for model convergence, each selected sample will be weighted by $\frac{1}{probability\times data\ size}$ to achieve {\small$\mathbb{E}_{(x,y)\sim P(\mathcal{S})}[f(x)]\!=\!\sum_{(x,y)\in\mathcal{S}}P(x)\cdot\frac{f(x)}{P(x)\cdot |\mathcal{S}|}=\mathbb{E}_{(x,y)\sim\mathcal{S}}[f(x)]$}.

\subsection{Proof of Lemma \ref{lemma: optimal distribution for minibatch sgd}}
\label{appendix: proof of Lemma 2}
According to our previous analysis, term $\beta_y$ for each class $y\!\in\!\mathcal{Y}$ is uniquely determined by its intra-class data selection strategy $P_{t,y}$ while term $\gamma_y$ is a fixed value. Therefore, we can minimize the overall gradient variance as follows:

        First, we derive the minimal $\beta_y^*$ by optimizing $P_{t,y}$, which can be directly solved using Cauchy-Schwarz inequality:
        \begin{equation}
            \begin{small}
            \begin{aligned}
                \beta_y\ge & \sum_{(x,y)\in\mathcal{S}_y}\frac{\big|\big|\nabla l(w_t,x,y)\big|\big|^2}{|\mathcal{S}_y|^2\cdot\frac{\big|\big|\nabla l(w_t,x,y)\big|\big|}{\sum_{(x',y')\in\mathcal{S}_y}\big|\big|\nabla l(w_t,x',y')\big|\big|}}\\
                = & \Big[\sum_{(x,y)\in\mathcal{S}_y}\frac{\big|\big|\nabla l(w_t,x,y)\big|\big|}{|\mathcal{S}_y|}\Big]^2,
            \end{aligned}
            \end{small}
            \nonumber
        \end{equation}
        where the equality holds when {\small $P_{t,y}(x)=\frac{\big|\big|\nabla l(w_t,x,y)\big|\big|}{\sum_{(x',y')\in\mathcal{S}_y}\big|\big|\nabla l(w_t,x',y')\big|\big|}$} based on the Cauchy-Schwarz inequality.
        
        Second, given $(\beta_y^*\!-\!\gamma_y)$ for each class, we minimize the overall objective $\sum_y\alpha_y(\beta_y^*\!-\!\gamma_y)$ by optimizing $|\mathcal{B}_y|$, the analytical expression of which can also be computed through Cauchy-Schwarz inequality:
        \begin{equation}
            \begin{small}
            \begin{aligned}
                \sum_{y\in\mathcal{Y}}\alpha_y\cdot(\beta_y^*-\gamma_y)
                = & \sum_{y\in \mathcal{Y}} \frac{|\mathcal{S}_y|^2}{|\mathcal{S}|^2\cdot |\mathcal{B}_y|}\cdot (\beta_y^*-\gamma_y)\\
                = & \frac{1}{|\mathcal{S}|^2}\cdot \sum_{y\in\mathcal{Y}}\frac{|\mathcal{S}_y|^2\cdot(\beta_y^*-\gamma_y)}{|\mathcal{B}_y|},
            \end{aligned}
            \end{small}
            \label{proof: lemma 2-2}
            \nonumber
        \end{equation}
        which is minimized when $\mathcal{B}_y\propto|\mathcal{S}_y|\sqrt{\beta_y^*-\gamma_y}$ according to the Cauchy-Schwarz inequality and leads to Lemma \ref{lemma: optimal distribution for minibatch sgd}.

\section{Supplementary Evaluation}
\label{appendix: extended experiments}
To further demonstrate the generality of {\sf Titan} framework, we also evaluate its performance in diverse practical scenarios to demonstrate the robustness and generality of {\sf Titan}, such as  fluctuant on-device idle computing resources, federated learning scenario and noisy on-device data. The experiments are mainly conducted on image classification task.

\begin{figure}
    \centering
    \includegraphics[width=0.7\linewidth]{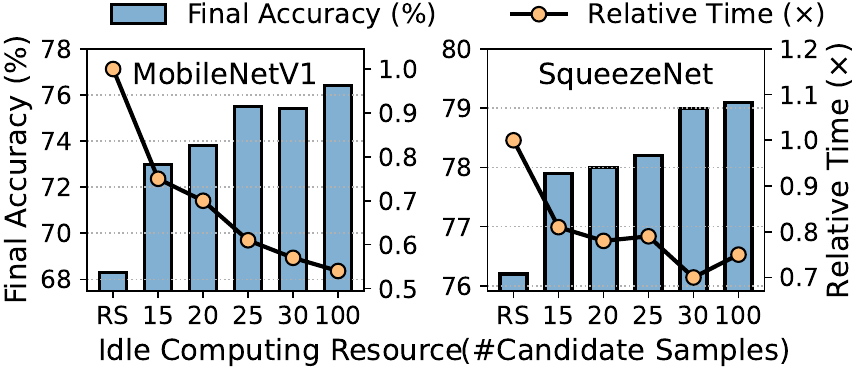}
    \caption{Performance on fluctuant idle computing resource.}
    \Description{Performance on fluctuant idle computing resource.}
    \label{fig: varying proportions of idle computational resources}
\end{figure}

\textbf{Fluctuant Idle Computing Resources.}
In practice, the co-running applications may occupy varied proportions of computing resources. 
When there exists more idle computing resource, a larger candidate dataset can be filtered out by coarse-grained filter to facilitate a higher-quality fine-grained data selection process.
Experimental results in Figure \ref{fig: varying proportions of idle computational resources} show that when the candidate dataset size rises from $15$ to $100$, the final accuracy of {\sf Titan} is increased from 
$73.0\!-\!77.9\%$ to $76.4\!-\!79.1\%$ and the training time reduction also rises from $19\!-\!25\%$ to $25\!-\!46\%$. The consistent improvement in model training performance demonstrates {\sf Titan}'s robustness to devices with varying idle computing resources.

\begin{figure}
    \centering
    \includegraphics[width=0.5\linewidth]{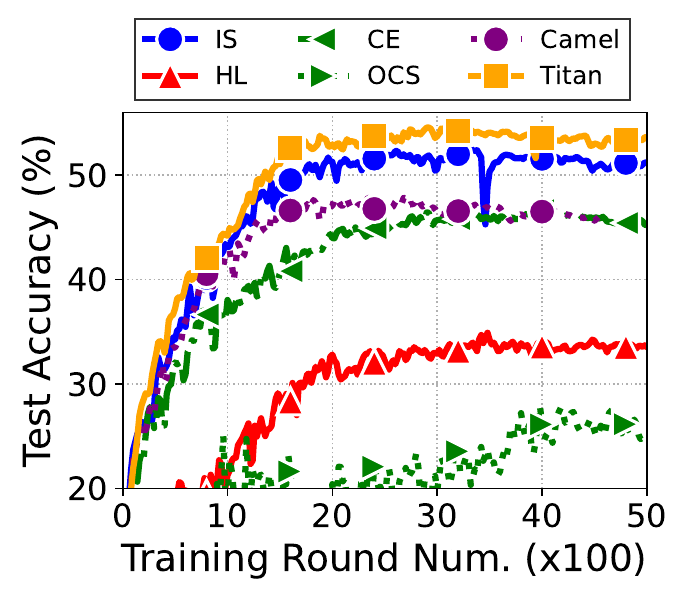}
    \caption{Performance in federated learning scenario.}
    \Description{Performance in federated learning scenario.}
    \label{fig: federated learning}
\end{figure}

\textbf{Federated Learning.} 
We evaluate the performance of {\sf Titan} in a federated setting setting using CIFAR-10 and MobileNetV1. In this scenario, we initialize the training process with $50$ devices, and the data distribution of across these devices follow a non-IID pattern as described in previous work~\cite{DBLP:conf/mlsys/BonawitzEGHIIKK19, tahir2022fedss}. The data on each device covers only $5$ classes. 
In each training round, random $20\%$ devices participated in the model training process, which independently select $|\mathcal{B}|$ samples from a pool of $v$ real-time samples, update their local models for 3 local training iterations, and upload the updated model to the centralized server for model aggregation.
As shown in Figure \ref{fig: federated learning}, compared with the second best best approach, {\sf Titan} achieves an increase of $2.03\%$ in final accuracy and speeds up the training time (i.e., number of communication rounds) to target accuracy by a factor of $3.17\times$. These findings highlight the potential of applying {\sf Titan} to federated learning for improving the training performance of global model.
\begin{figure}
    \subfigure[Feature Noise Setting]{
        \includegraphics[width=0.465\linewidth]{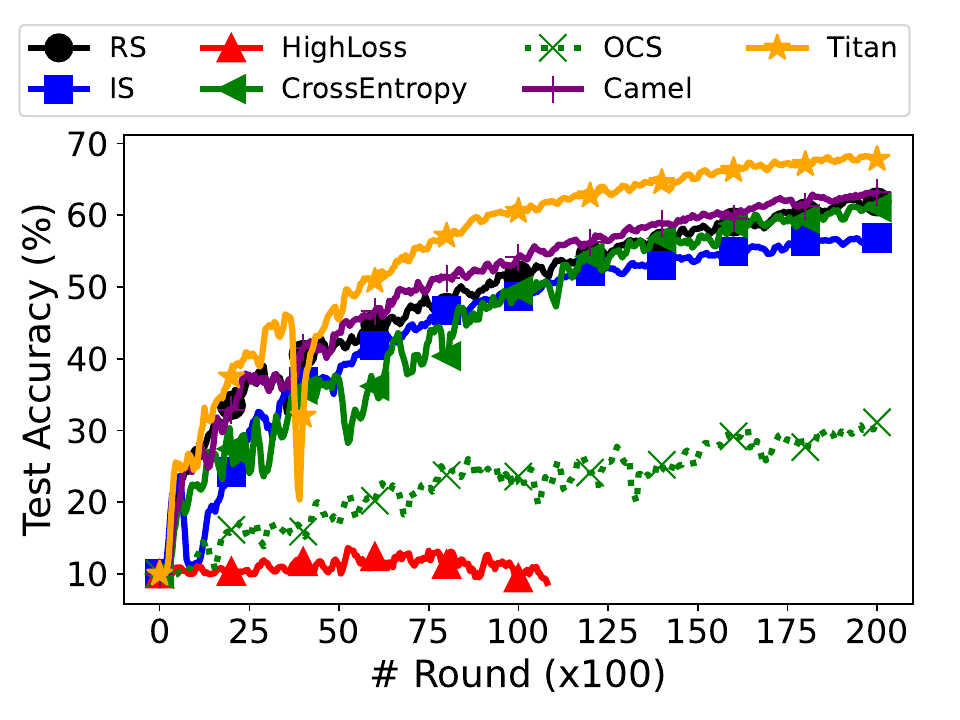}
    }
    \subfigure[Label Noise Setting]{
        \includegraphics[width=0.465\linewidth]{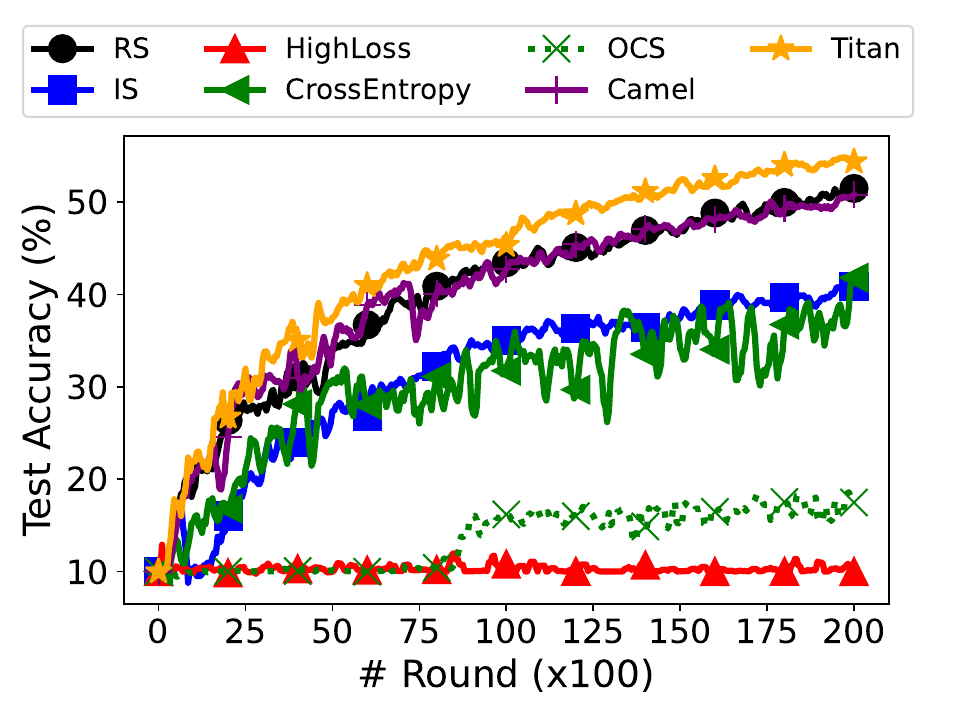}
    }
    \caption{{\sf Titan} Performance on noisy data distribution.}
    \Description{{\sf Titan} Performance on noisy data distribution.}
    \label{fig: noise}
\end{figure}

\textbf{Noisy Data Streams.}
Generally, {\sf Titan} framework can improve the model training performance for arbitrary on-device data distribution, as the importance of each data sample in our work is measured by its theoretical contribution to model performance over on-device personal data (elaborated in Theorem \ref{theorem: convergence rate} and Lemma \ref{lemma: optimal distribution for minibatch sgd}). 
Specifically, for streaming data with noisy or shifted data distributions, {\sf Titan} aims to minimize the discrepancy between the optimal model (for noisy or dynamic data distribution) and current model (trained with the selected training data batch, which essentially reflect the model's robustness to noisy data and generalization performance to domain shift. 
Furthermore, we conduct extensive experiments to demonstrate the applicability of {\sf Titan} framework to noisy and dynamic data streams.

For noisy data, we incorporate two distinct type of noise:
(i) feature noise to emulate the noise introduced by an unstable environment during data collection, where we randomly select $40\%$ of data samples and add Gaussian noise to their input feature $x$;
(ii) label noise to simulate the incorrect labels arising from automatic-labeling where we randomly revise the label $y$ of $40\%$ data samples.
Experiment results on CIFAR-10 with MobileNetV1 are presented in Figure \ref{fig: noise}, which demonstrate that:
(i) In various noise settings, {\sf Titan} consistently outperforms all the baselines, achieving $6\%$ and $3.4\%$ higher final model accuracy in the settings of feature noise and label noise.
Moreover, {\sf Titan} accelerates the time-to-accuracy by a factor of $1.83\times$ and $1.35\times$ in these two settings;
(ii) {\sf Titan} exhibits higher robustness to feature noise compared to label noise, because label noise introduces larger errors to the sample gradients and leads to inaccurate data evaluation.